\documentclass[letterpaper,journal]{IEEEtran}

\usepackage{iftex}
\ifPDFTeX
  \usepackage[T1]{fontenc}
\fi
\usepackage{amsmath,amsfonts,amssymb}
\usepackage{newtxtext,newtxmath}
\usepackage{algorithmic}
\usepackage{algorithm}
\usepackage{array}
\usepackage{booktabs}
\usepackage{tabularx}
\usepackage[caption=false,font=normalsize,labelfont=sf,textfont=sf]{subfig}
\usepackage{textcomp}
\usepackage{stfloats}
\usepackage{graphicx}
\usepackage{cite}
\usepackage{placeins}
\usepackage[hidelinks]{hyperref}
\usepackage{orcidlink}

\graphicspath{{figures_v6/}}

\newcommand{\dk}{\ensuremath{\Delta\kappa}}
\newcommand{\msd}[2]{#1\,$\pm$\,#2}
\begin{document}

\title{NeuralParker: A Reinforcement Learning Planner for
Irregular Parking Environments}

\author{Zihan~Wang~\orcidlink{0009-0004-5852-9844},
        Bai~Huang~\orcidlink{0000-0001-6550-4427},
        Yang~Guan~\orcidlink{0000-0003-0689-0510},
        Xiao~Li,
        Haoyu~Xu,
        Naizheng~Wang,
        and~Shengbo~Eben~Li~\orcidlink{0000-0003-4923-3633}%
\thanks{This work was supported in part by the Tsinghua University--Meituan
Joint Institute for Digital Life under Grant 20252930041, and in part by the
National Natural Science Foundation of China under Grant 92582205. This work
was completed during Zihan Wang's internship at Meituan. (\textit{Zihan Wang
and Bai Huang contributed equally to this work.}) (\textit{Corresponding
author: Yang Guan.})}
\thanks{Zihan Wang is with the School of Life Sciences, Tsinghua University,
Beijing 100084, China (e-mail: zihanwan24@mails.tsinghua.edu.cn).}
\thanks{Bai Huang is with the School of Statistics and Mathematics, Central
University of Finance and Economics, Beijing 100081, China (e-mail:
huangbai@cufe.edu.cn).}
\thanks{Yang Guan and Shengbo Eben Li are with the School of Vehicle and
Mobility, Tsinghua University, Beijing 100084, China (e-mail:
yguan@tsinghua.edu.cn; lishbo@tsinghua.edu.cn).}
\thanks{Xiao Li, Haoyu Xu, and Naizheng Wang are with Meituan Autonomous
Delivery, Beijing 100102, China (e-mail: lixiao07@meituan.com;
xuhaoyu06@meituan.com; wangnaizheng@meituan.com).}}

\markboth{Preprint. Submitted to IEEE Transactions on Intelligent Transportation Systems}%
{Wang \MakeLowercase{\textit{et al.}}: NeuralParker for Arbitrary-Pose Parking}

\maketitle

\begin{abstract}
Automated parking commonly assumes marked slots and short approach maneuvers.
Delivery and service vehicles, however, may need to reach an operator-specified
pose in an irregular bounded environment from a distant start. Existing
learning-based parking planners often rely on local observations, which can
restrict long-range route reasoning. To address this problem, we present
NeuralParker, a reinforcement learning-based hybrid planner for arbitrary-pose
parking. NeuralParker encodes full-environment
obstacle and boundary geometry in a target-relative vertex representation,
allowing the policy to retain route-defining context throughout the approach.
It further couples a learned curvature--length arc policy with an in-loop
terminal ensemble that selects from diverse cubic Hermite connections using a
curvature-regularized cost. We also establish factorial and
long-range route-choice benchmarks to evaluate planning success and trajectory
quality. Experiments on these benchmarks show that NeuralParker achieves higher
planning success and better overall trajectory quality than the evaluated
baselines, while ablation studies support the benefits of the target-relative
global representation and terminal ensemble. Finally, a real-vehicle
evaluation confirms that the planner transfers effectively to real
delivery-vehicle perception at a working parking site, planning successfully at
low computational cost.
\end{abstract}


\section{Introduction}

\IEEEPARstart{A}{utomated} parking is a representative low-speed
motion-planning task for autonomous vehicles~\cite{paden2016,banzhaf2017}.
Most formulations ask a nonholonomic
vehicle to enter a marked parallel or perpendicular slot from a nearby aisle.
Freight, delivery, and service vehicles face a broader requirement: an
infrastructure map, task planner, or operator may specify an exact pose beside
a pallet, loading point, or service entrance~\cite{freightpose2021}. The
vehicle may begin far from this target, and the drivable region may contain
nonconvex boundaries, pockets, corridors, and irregularly placed obstacles. We
refer to this setting as \emph{arbitrary-pose parking in an irregular
environment}: the goal is task-defined rather than slot-defined, and planning
must cover both the approach and the final maneuver.

Existing parking planners broadly fall into two groups: traditional
model-based planners and planners with learned components. Within the latter,
hybrid systems retain model-based geometric stages. Traditional planners
retain explicit vehicle kinematics and environment geometry throughout each
query. Curvature-bounded connectors provide direct pose-to-pose
connections~\cite{dubins1957,reedsshepp1990}; state-lattice and Hybrid A*
search extend geometric reasoning to obstacle-rich
spaces~\cite{pivtoraiko2009,dolgov2010}; and constrained optimization enforces
kinematic and collision constraints~\cite{optca2018,optiter2021}.
Hierarchical systems further decompose long-horizon parking into route
planning, local maneuvering, and trajectory repair~\cite{dai2021}. These
methods provide explicit map-based feasibility reasoning and remain strong
geometric baselines. For distant starts in irregular environments, they solve a
search, trajectory-repair, or nonlinear optimization problem at query time.
Planners with learned components instead amortize maneuver selection over a
training distribution.

Planners with learned components differ in both training paradigm and planning
architecture. Direct reinforcement-learning (RL) policies learn maneuvers from
compact states or local range
observations~\cite{rle2e2019,rlmp2020,parkbench2026}, whereas imitation-learning
systems predict waypoints or motion segments from camera or bird's-eye-view
features~\cite{parkinge2e2024,parkformer2025,multipark2025}. At the
architectural level, hybrid planners retain geometric stages: HOPE and RL-OGM
combine a learned local planner with analytic completion, whereas N3P predicts
an intermediate pose for subsequent geometric
planning~\cite{hope2024,rlogm2025,n3p2026}.
Despite this progress, three design and evaluation issues are central to the
setting considered here. First, compact, ego-centered, or finite-range
observations omit target-side geometry before a long-range approach is chosen,
so the policy needs full-environment context alongside near-field clearance
sensing. Second, expanding an analytic connector changes the approach states
from which a terminal connection can be found and accepted, and its interaction
with policy training calls for controlled evidence. Third, representative
evaluations cover variations in slot type, clearance, obstacle layout, or
spatial confinement~\cite{parkbench2026,hope2024,drip2025}, so long-range route
choice calls for tests that separate full-set success rate from common-success
trajectory quality. Together, these issues motivate a coordinated treatment of
full-environment representation, terminal-connector design, and benchmark
construction for distant-start arbitrary-pose parking.

We address these issues with \emph{NeuralParker}. The target defines a common
coordinate frame in which polygonal obstacles, circular obstacles, and
environment boundaries are represented as vertices within a fixed object
budget; a stream of footprint-rectified local LiDAR clearance rays computed
from the same geometry provides near-field input. An attention-based policy maps these observations to
curvature--length arcs. After each collision-free, nonterminal learned arc
that satisfies a target-frame heading gate, an in-loop cubic-Hermite ensemble
varies terminal-pose hypotheses and both endpoint tangent scales, rejects
candidates that violate the sampled geometry or curvature constraints, and
selects a low-cost feasible connection. Training and evaluation use the same
feasibility, selection, and terminal-cost rules.

The contributions are:
\begin{itemize}
\item We introduce a target-relative vertex representation that augments local
clearance rays with full-environment obstacle and boundary geometry. Against a
local-only observation and a pretrained bird's-eye-view raster, it gives the
best success rate and path quality under two different policy backbones, and needs
no image rendering or autoencoder pretraining.
\item We introduce a curvature-regularized Hermite terminal ensemble that
replaces a fixed connector with a diverse candidate set under a shared
curvature-aware cost. It improves success rate and path quality when substituted at
evaluation time alone, and the gains persist when the policy is trained through
the same rule.
\item We establish factorial and long-range route-choice benchmarks that
separate full-set success rate from common-success trajectory quality, and
report that NeuralParker attains the highest success rate on both benchmarks
together with the shortest and smoothest trajectories among the compared
planners.
\end{itemize}

The remainder of this article is organized as follows.
Section~\ref{sec:related} reviews traditional and learning-based parking
planners. Section~\ref{sec:method} presents the observation, arc policy,
terminal ensemble, and training objective. Section~\ref{sec:setup} describes
the benchmarks, baselines, and metrics, and Section~\ref{sec:results} reports
the full-system comparison, the two ablations, and the real-vehicle
experiment. Section~\ref{sec:conclusion} concludes the article.

\section{Related Work}
\label{sec:related}

We review parking planners along the two axes that this work couples: how the
environment is represented, and how the final pose is reached. Traditional
planners are considered first, then learning-based and hybrid planners.

\subsection{Traditional Parking Planners}

Traditional planners model vehicle kinematics and environment geometry
explicitly. Dubins and Reeds--Shepp curves solve curvature-bounded pose
connections, with reverse motion available in the latter; parking-specific
arc--clothoid and parametric constructions adapt this idea to local
maneuvers~\cite{dubins1957,reedsshepp1990,vorobieva2015,upadhyay2018}. These connectors
are efficient when a direct boundary-value solution is feasible, but they
require global planning when obstacles block it.

Global planners provide this reasoning through search, sampling, or
optimization~\cite{paden2016,banzhaf2017}. State lattices and Hybrid A*
combine motion primitives with analytic expansion, and multiresolution or
guided variants reduce search effort in constrained parking
environments~\cite{pivtoraiko2009,dolgov2010,tazaki2017,guidedha2019,fastastar2021}.
RRT-based planners explore continuous free space~\cite{rrtpark2018,rrtavp2021},
whereas optimization methods enforce kinematics and collision constraints
through dynamic formulations, dual variables, or iteratively constructed
corridors~\cite{li2015unified,optca2018,optiter2021}. Hierarchical systems
further select preparatory poses or couple long-horizon routing with local
repair~\cite{reachable2023,dai2021}. These methods retain map-wide geometry
and checked feasibility; NeuralParker keeps both properties in a learned policy
that amortizes the repeated planning decisions.

\subsection{Learning-Based Parking Planners}

Learning-based parking planners replace part of this geometric pipeline with a
trained model. Deep-RL methods use compact target-relative states, local range or
bird's-eye-view (BEV) observations, and demonstrations to produce parking
actions, including hierarchical, federated, and action-chunked
variants~\cite{rle2e2019,rlmp2020,drltraj2020,revpark2019,fedpark2023,rlexpert2023,
humanrev2024,parkbench2026}. Imitation-learning systems instead predict
waypoints or motion segments from surround-view or BEV
features~\cite{parkinge2e2024,parkformer2025,transparking2025,multipark2025}. These
methods support direct maneuver generation but typically bind useful spatial
context to a compact state, sensor horizon, local raster, or training
trajectories. Structured vector encoders capture map elements for motion
prediction and vectorized driving
policies~\cite{vectornet2020,vad2023}; NeuralParker applies related structured
geometry to target-conditioned parking control.

Hybrid planners retain analytic stages. HOPE fuses finite-range obstacle
distances, an ego-centered BEV crop, a relative target, and an action mask,
and switches between learned increments and feasible Reeds--Shepp completion
during training~\cite{hope2024}. RL-OGM similarly combines a local LiDAR
occupancy grid with Reeds--Shepp planning; DRIP diffusion-refines an RL prior
under spatial constraints; and N3P predicts a preparatory pose before Hybrid
A* and Reeds--Shepp stages~\cite{rlogm2025,drip2025,n3p2026}. These systems
motivate learned--analytic coupling. We isolate the effect of expanding
the analytic connector under both evaluation-only and matched
train-and-evaluate interventions.
HOPE, RL-OGM, and DRIP retain local obstacle context, whereas N3P makes the
global-to-local handoff an explicit staged decision. NeuralParker therefore
studies full-environment target-relative geometry and connector--policy
coupling as separate factors.

Beyond parking, RL also supports broader automated-driving decision and control.
Integrated decision and control (IDC) trains an approximate constrained
optimal-control-problem (OCP) solver offline for online path selection and
tracking~\cite{guan2023idc}, and its enhanced formulation combines experience
data with an analytic model, incorporates attention-based road-user encoding,
and was evaluated on an automated vehicle at a signalized
intersection~\cite{guan2026enhancedidc}. These systems share NeuralParker's
premise that a learned component can absorb repeated online optimization,
applied there to structured on-road traffic and here to pose-defined parking in
an irregular bounded region.

\section{Methodology}
\label{sec:method}

\subsection{Problem Definition and Evaluation Protocol}

We model parking as a Markov decision process (MDP)
$(\mathcal{S},\mathcal{A},p,r)$ over a scenario
distribution $\Omega$. A scenario contains an ego pose, a target pose, static
obstacles, and a bounded drivable region. The planner must produce a
reference-point trace that passes the sampled obstacle and boundary checks and
terminates within $0.1$ m and $10^\circ$ of the target.
The policy is optimized with proximal policy optimization (PPO), a direct
policy-gradient method based on a
clipped surrogate objective~\cite{guan2021directindirect,ppo2017}:
\begin{equation}
\arg\max_\theta\;\mathbb{E}\!\left[
\min\!\left(\rho_\theta A,
\mathrm{clip}(\rho_\theta,1-\epsilon,1+\epsilon)A\right)\right],
\label{eq:ppo}
\end{equation}
where $\rho_\theta=\pi_\theta(a|o)/\pi_{\theta^-}(a|o)$. The policy observation
$o$ is specified below; the environment additionally retains the preceding
action to evaluate curvature and gear-change costs.

We use one fixed full-range start protocol throughout. For each benchmark,
initial poses are drawn once from the environment's full-range sampler without
distance rejection or banding, using sampling seed 0. We freeze 40
target-relative poses per test scene and replay exactly the same start sets for
every method and policy-training seed: 1080 episodes for the 27-scene
Factorial Parking Benchmark and 720 for the 18-scene Topology-Stress
Benchmark. The sampling seed only identifies this frozen evaluation set and is independent
of the policy-training seeds. Success is computed on every episode. In every
comparative table, path length, full-trajectory reversals, and accumulated
absolute curvature change are computed on one paired cohort: the episodes
solved by \emph{all} methods in that table. Method-specific own-success
statistics are retained only as diagnostics and are not used as cross-method
evidence.

\subsection{Target-Relative Full-Environment Observation}

The policy receives
\begin{equation*}
o=(s^e,s^p,s^o,s^b,s^\ell),
\end{equation*}
where $s^e$ is the ego state, $s^p$ polygonal-obstacle vertices, $s^o$
circular obstacles represented as octagons, $s^b$ boundary segments, and
$s^\ell$ an optional local LiDAR stream. The reported experiments use 120
uniformly spaced ego-centric beams over $360^\circ$. Let $d_i$ denote the
reference-point distance to the first obstacle or boundary intersection,
clipped at 10 m, and let $b_i$ denote the direction-dependent distance from
the reference point to the boundary of the rectangular vehicle footprint
adopted from HOPE. The input to NeuralParker is the rectified clearance
\begin{equation}
s^\ell_i=\max(d_i-b_i,0).
\label{eq:lidar-clearance}
\end{equation}
The ray-wise footprint correction in~\eqref{eq:lidar-clearance} is an
observation convention; episode termination still uses the sampled
reference-point collision test defined above. Appendices~\ref{app:implementation} and~\ref{app:hope-adaptation} give
the remaining NeuralParker and adapted-HOPE beam conventions, respectively.
All other geometry is expressed in a target-relative frame with the target at
the origin and its heading aligned with the positive $y$-axis. Thus, unlike a
range-limited crop, target-side geometry remains represented regardless of the
ego--target separation.

Let the target pose in the world frame be $(p^g,\theta^g)$. A point $q$ and
heading $\theta$ are transformed as
\begin{equation}
\tilde q=R\!\left(\frac{\pi}{2}-\theta^g\right)(q-p^g),\qquad
\tilde\theta=\mathrm{wrap}\!\left(\theta-\theta^g+\frac{\pi}{2}\right),
\label{eq:target-frame}
\end{equation}
so the same representation is invariant to translating or rotating the whole
scene. Coordinates are not clipped by ego--target distance.

We apply~\eqref{eq:target-frame} to every stored vertex and
segment endpoint, so all geometry shares one metric frame.
We store variable-size sets in fixed-capacity slots, fill unused slots with a
sentinel value, and give every slot a learned positional embedding. The
scene encoder applies self-attention to these fixed slots
without passing an explicit padding mask to the attention layer. The
object budget is 20 and overflow raises an error instead of silently discarding
geometry.

\subsection{Arc Policy and Scene Encoder}

The policy outputs one curvature--length arc,
\begin{equation*}
a_t=(\kappa_t,l_t),\qquad \kappa_t\in[-0.68,0.68],\quad l_t\in[-10,10],
\end{equation*}
where the sign of $l_t$ selects forward or reverse motion. For nonzero
curvature, $r=1/|\kappa_t|$, $\alpha=|l_t\kappa_t|$, and the deterministic
transition is
\begin{align}
x^e_{t+1}&=x^e_t+\delta_l\!\left[r(1-\cos\alpha)\delta_\kappa\sin\theta^e_t
+r\sin\alpha\cos\theta^e_t\right], \nonumber\\
y^e_{t+1}&=y^e_t+\delta_l\!\left[-r(1-\cos\alpha)\delta_\kappa\cos\theta^e_t
+r\sin\alpha\sin\theta^e_t\right], \label{eq:arc}\\
\theta^e_{t+1}&=\theta^e_t-\delta_\kappa\alpha, \nonumber
\end{align}
with $\delta_l=\mathrm{sgn}(l_t)$ and
$\delta_\kappa=\mathrm{sgn}(\kappa_t)$. Under this simulator convention, the
signed length reverses the displacement, while the heading increment is
independent of $\delta_l$.

\begin{figure*}[!t]
\centering
\includegraphics[width=0.96\textwidth,trim=0 0 0 14bp,clip]{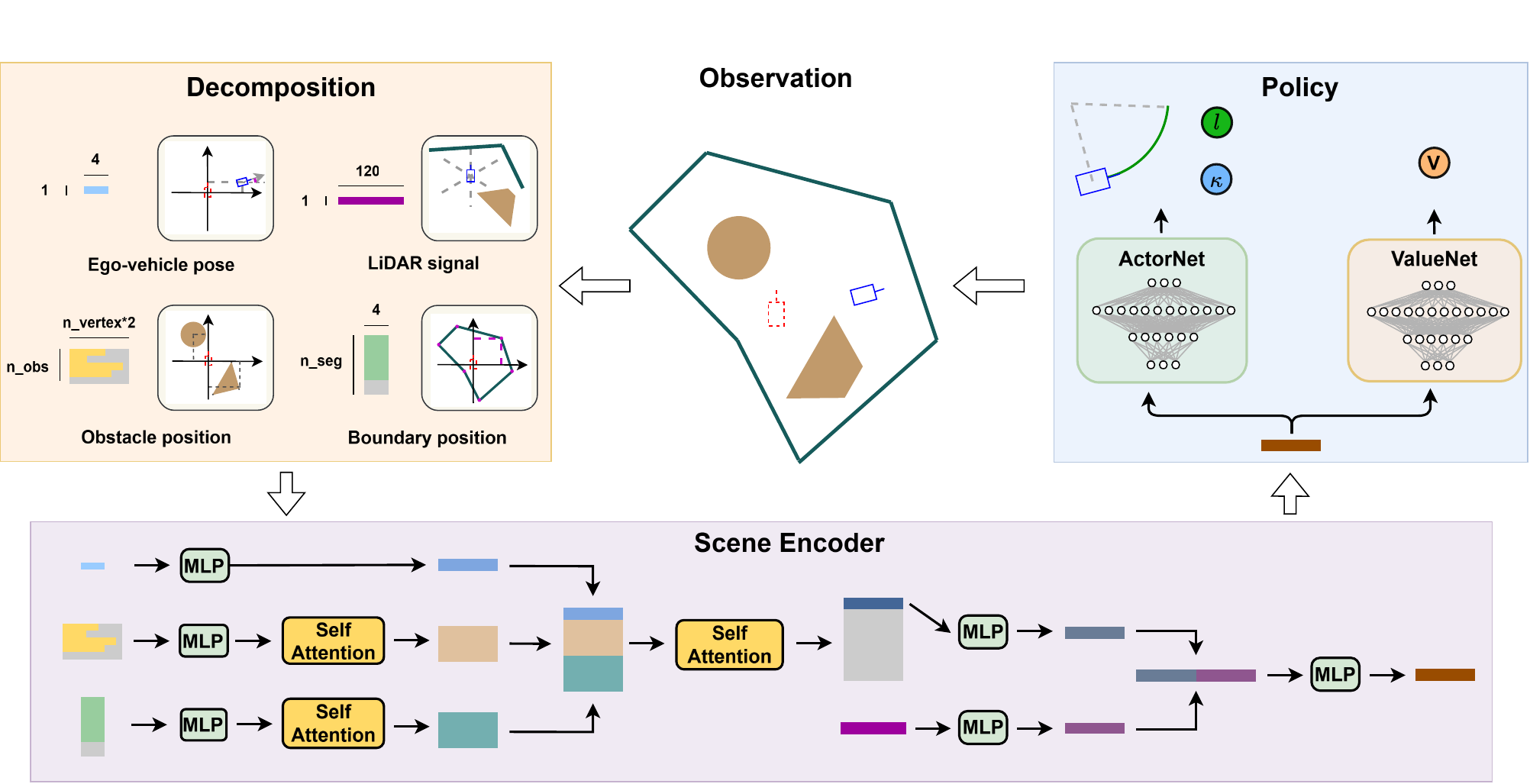}
\caption{Overview of the NeuralParker architecture. The observation is
decomposed into the ego pose, obstacle vertices, boundary segments, and local
LiDAR signals. The scene encoder embeds the four streams, uses self-attention
to aggregate the obstacle and boundary tokens, and fuses the resulting global
geometry with the ego and LiDAR features. The shared representation is passed
to the actor and critic networks, which output the curvature--length arc
$(\kappa,l)$ and the state-value estimate, respectively.}
\label{fig:architecture}
\end{figure*}

The scene encoder shown in Fig.~\ref{fig:architecture} embeds the ego,
global-geometry, and optional local-range streams with stream-specific MLPs. Geometry tokens receive positional
embeddings and are processed by self-attention~\cite{vaswani2017}; the fused
representation feeds actor and critic heads.
For token matrix $Z$, one attention layer computes
\begin{equation*}
\mathrm{Attn}(Z)=\mathrm{softmax}\!\left(
\frac{(ZW_Q)(ZW_K)^\top}{\sqrt{d}}\right)ZW_V,
\end{equation*}
followed by a feed-forward block and residual normalization. The actor
parameterizes a diagonal Gaussian over the normalized two-dimensional action,
while the critic estimates $V_\psi(o_t)$.

\subsection{In-Loop Hermite Terminal Ensemble}
\label{sec:hermite-method}

After a learned arc has been executed, collision checked, and tested for direct
terminal success, the planner applies the Hermite connector only when the
reached target-frame heading satisfies $\sin\tilde\theta>0$. Within this gate,
it tries a cubic Hermite connection
\begin{equation}
\begin{aligned}
g(t)=&(2t^3-3t^2+1)g(0)+(3t^2-2t^3)g(1)\\
&+(t^3-2t^2+t)g'(0)+(t^3-t^2)g'(1).
\end{aligned}
\label{eq:hermite}
\end{equation}
In~\eqref{eq:hermite}, $g(0)$ and $g(1)$ are the reached and terminal
positions and $g'(0)$ and $g'(1)$ the corresponding tangents.
The unscaled tangent magnitude equals the Euclidean distance between the two
endpoints; the scale factors below multiply this magnitude independently at
the two endpoints.
Within the gate, the connector uses forward tangents when the reached
target-frame position satisfies $\tilde y<0$ and reverse tangents otherwise.
For the selected direction, the target endpoint is perturbed by three lateral
offsets and three heading offsets, producing nine baseline candidates.
Candidates are rejected if they collide or violate the curvature bound.

The ensemble additionally varies start and target tangent scales. The default
independent grid uses scales $\{0.8,1.0,1.2\}$ at both endpoints, giving
$9\times3\times3=81$ candidates. The reported implementation evaluates
$N=100$ uniformly sampled points $g_j=g(t_j)$, with
$\Delta t=1/(N-1)$, and computes
\begin{align}
v_j&=(g_{j+1}-g_j)/\Delta t,\qquad
\hat v_j=v_j/\|v_j\|, \nonumber\\
\bar\kappa_j&=
\frac{\|\hat v_{j+1}-\hat v_j\|}{\Delta t\,\|v_j\|}, \nonumber\\
\Delta\kappa_H&=|\kappa_0-\bar\kappa_0|
+\sum_{j=1}^{N-3}|\bar\kappa_j-\bar\kappa_{j-1}|,
\label{eq:hermite-kappa}
\end{align}
where $j=0,\ldots,N-3$, $\bar\kappa_j\geq0$ is the sampled curvature
magnitude, and $\kappa_0$ is the signed curvature of the preceding action arc.
This discrete convention is used by all evaluated policies. A
candidate is feasible only if all sampled points are inside the boundary,
outside every obstacle, and $\max_j\bar\kappa_j\leq0.68$. Among feasible
candidates, the implementation selects
\begin{equation}
i^*=\arg\min_i\left(10\Delta\kappa_{H,i}+10L_{H,i}+5\right).
\label{eq:mincost}
\end{equation}
The terminal reward instead weights curvature by $0.05$ and clips the
ensemble score as defined below. The fixed nine-candidate connector accepts
the first feasible curve and retains the corresponding unclipped score.
Consequently, the evaluation-only configuration measures test-time connector
replacement, whereas the train-and-evaluate configuration measures end-to-end
coupling to the ensemble and its terminal-cost rule.

\subsection{Reward and Training}
\label{sec:reward}

The reported experiments use the path-quality reward defined below. For each
policy arc, before any terminal addend, the implemented reward is
\begin{equation}
\begin{split}
r_t^{\mathrm{step}}={}&-0.05|\kappa_t-\kappa_{t-1}|-10|l_t|-5\\
&-100\,\mathbb{I}[l_tl_{t-1}<0]
-2\,[\max(0,0.5-|l_t|)]^2 .
\end{split}
\label{eq:step-reward}
\end{equation}
The last term in~\eqref{eq:step-reward} discourages near-zero-length actions and
the associated sign jitter. For the first arc, both history-dependent terms are
set to zero. Let
$N_{\mathrm{rev}}$ be the number of sign changes among policy arcs
and
\begin{equation}
\Delta\kappa_{\mathrm{eff}}=
\sum_{t=2}^{T}\left|\mathrm{sgn}(l_t)\kappa_t-
\mathrm{sgn}(l_{t-1})\kappa_{t-1}\right|.
\label{eq:effective-kappa}
\end{equation}
On success, the environment adds
\begin{equation}
\begin{aligned}
r^{\mathrm{succ}}&=1250.5-20N_{\mathrm{rev}}
-\Delta\kappa_{\mathrm{eff}}+1.5p_H,\\
p_H&=\max\!\left\{
-\left(0.05\Delta\kappa_H+10L_H+5\right),-500\right\},
\end{aligned}
\label{eq:terminal-reward}
\end{equation}
where $\Delta\kappa_{\mathrm{eff}}$ is given by
\eqref{eq:effective-kappa}, $p_H=0$ for a direct terminal state, and
$(L_H,\Delta\kappa_H)$ are the selected Hermite segment's length and discrete
curvature variation. The
fixed nine-candidate configuration uses the same expression without the $-500$
floor, which bounds the terminal penalty in the ensemble implementation.
Collision or boundary failure adds $-2501$; timeout receives no success bonus.
Equation~\eqref{eq:mincost} uses the larger curvature weight only for candidate
ranking, whereas~\eqref{eq:terminal-reward} defines the terminal learning
signal.

NeuralParker uses an obstacle-aware reverse start curriculum. Initial
states are ordered by a 1.5 m-grid geodesic distance, beginning at 5\% of the
available range and expanding during
training~\cite{florensa2017,barc2019}; the matched control in
Section~\ref{sec:curriculum-results} disables only this ordering. The
adapted HOPE baseline in the main comparison is trained without a start-state
curriculum.

Algorithm~\ref{alg:neuralparker} summarizes the coupled training loop. The
important distinction from a post-processing spline is that the same Hermite
feasibility test determines successful transitions in the rollout data used by
PPO.

\begin{algorithm}[!t]
\caption{NeuralParker training with an in-loop Hermite ensemble}
\label{alg:neuralparker}
\begin{algorithmic}[1]
\REQUIRE scene distribution $\Omega$, policy $\pi_\theta$, value $V_\psi$
\FOR{each PPO collection iteration}
  \STATE sample a scene and curriculum start; observe $o_0$
  \WHILE{the episode is active}
    \STATE sample $(\kappa_t,l_t)\sim\pi_\theta(\cdot\mid o_t)$
    \STATE execute the arc; compute $r_t^{\mathrm{step}}$ and set $d_t\gets0$
    \IF{the executed arc is invalid}
      \STATE add the failure penalty; set $d_t\gets1$
    \ELSIF{the reached state is directly terminal}
      \STATE add \eqref{eq:terminal-reward} with $p_H=0$; set $d_t\gets1$
    \ELSIF{$\sin\tilde\theta_{t+1}>0$}
      \STATE form feasible set $\mathcal F_t$ from the 81 candidates
      \IF{$\mathcal F_t\neq\varnothing$}
        \STATE select $i^*$ by \eqref{eq:mincost}
        \STATE add \eqref{eq:terminal-reward}; set $d_t\gets1$
      \ENDIF
    \ENDIF
    \STATE store $(o_t,a_t,r_t,o_{t+1},d_t)$
    \STATE set $o_t\gets o_{t+1}$
  \ENDWHILE
  \STATE update $(\theta,\psi)$ with the clipped PPO objective in \eqref{eq:ppo}
\ENDFOR
\end{algorithmic}
\end{algorithm}

\section{Experimental Setup}
\label{sec:setup}

\subsection{Scenario Design}

We construct two complementary procedurally generated benchmarks. The
\emph{Factorial Parking Benchmark} contains 81 training scenes and 27
structurally disjoint test scenes. Both splits follow the same
$3\times3\times3$ factorial design: three start-distance regimes (near,
medium, long), three parking types (reverse, lateral, random), and three
difficulty levels (easy, medium, hard). The training split contains three
structurally distinct scenes in every cell, while the test split contains one
additional scene per cell. Each evaluation seed runs 40 starts per
test scene, giving 1080 episodes.

The factorial design covers breadth, and many of its scenes are solvable from
local clearance and analytic completion. The \emph{Topology-Stress Benchmark}
targets long-range approach decisions, where target-side structure initially
lies beyond the 10 m local LiDAR view. It crosses three route-choice families. In
\emph{Type 1}, a roof over the target entrance and a wall on one side leave only
the opposite-side swing feasible. In \emph{Type 2}, a three-sided pocket opens
away from the nominal arrival direction, so a direct targetward approach reaches
the closed back wall and must detour to the entrance. In \emph{Type 3}, an outer
barrier blocks the direct approach and leaves a laterally displaced passage. In
each family, a range-limited
policy can favor an unfavorable approach before the decisive opening becomes
locally visible. The benchmark contains 48 training scenes
and 18 structurally disjoint test scenes, crossing these three families with
medium and hard difficulty. Each evaluation seed runs 40 shared starts per
test scene, giving 720 episodes. Across both benchmarks, all scenes pass the
same target-validity, object-capacity, connected-start, and analytic-feasibility
checks.

\subsection{Baselines}
\label{sec:baselines}

HOPE~\cite{hope2024} is the principal learned baseline because it is among the
strongest recent learning-based parking planners reported for diverse
scenarios. It also provides direct counterparts to both components studied
here: an ego-centered BEV raster augments its local range and target features,
and a shortest-first Reeds--Shepp controller takes over near the target. We make
three task-side changes for a fair comparison under a common task definition.
Specifically, we replace HOPE's native collision-and-retreat behavior with the
hard reference-point collision and boundary termination used by NeuralParker; we
check its Reeds--Shepp candidates under the benchmark's reference-point
collision and boundary convention; and we attach the fixed nine-candidate
Hermite terminator used by the controlled baselines. The complete adapted
baseline retains HOPE's action mask and RS
controller. Appendix~\ref{app:hope-adaptation} records the retained components,
task-side changes, and configurations used for the controlled HOPE-style
variants.

The real-vehicle study in Section~\ref{sec:realworld} adds two further
baselines: Hybrid A*~\cite{dolgov2010} and \emph{NeuralParker-Guided A*}. The
latter is a hybrid of the two, and keeps the Hybrid A* search complete while
using the learned policy to bias where that search looks. Concretely, the
NeuralParker policy is rolled out once from the query state to obtain a
reference path, and the distance from a lattice node to that path enters the
node cost as an additional term, so expansions that follow the learned
maneuver are explored first. Because the underlying search and its feasibility
tests are unchanged, this coupling alters only the expansion order: it can
reach goals that require an initially target-averse maneuver within the time
budget, and it reduces the number of expanded nodes when both planners agree on
the maneuver.

\subsection{Metrics and Training}

We report full-set success rate, path length, full-trajectory reversals, and
cumulative curvature change. Let
$\{(\kappa_t,l_t)\}_{t=1}^{T}$ denote the executed non-Hermite action arcs,
whether policy-produced or RS-controlled. When Hermite completion is used,
let $L_H$, $\Delta\kappa_H$, and $q_H\in\{-1,1\}$ denote its sampled length,
curvature variation from~\eqref{eq:hermite-kappa}, and drive direction.
The reported metrics are
\begin{align}
L&=\sum_{t=1}^{T}|l_t|+L_H, \nonumber\\
N_{\mathrm{rev}}&=\sum_{t=2}^{T}\mathbb I[l_tl_{t-1}<0]
+\mathbb I[l_Tq_H<0], \label{eq:metrics}\\
\dk&=\sum_{t=2}^{T}|\kappa_t-\kappa_{t-1}|+\Delta\kappa_H. \nonumber
\end{align}
For direct terminal success, the Hermite terms and final direction indicator
are omitted.
Every controlled simulation path-quality comparison uses the intersection of
successful episode identifiers across \emph{all} methods in that table;
success itself uses the full set. NeuralParker-backbone policies are trained with PPO for 240
epochs, whereas adapted-HOPE and HOPE-style policies use the HOPE PPO
implementation for 10,000 episodes. Appendices~\ref{app:implementation}
and~\ref{app:hope-adaptation} report the corresponding training configurations.
Results are three-seed mean $\pm$ sample standard deviation. In the controlled
simulation tables, boldface identifies
the best mean within a comparison and does not denote a
statistical-significance test. Quality values formed from different
table-specific common-success intersections are not compared across tables.

\section{Results}
\label{sec:results}

We organize the evaluation from planner-level comparisons to controlled
component ablations and finally to operation with real-world perception.
In the controlled simulation comparisons, success is evaluated on the full
test set, whereas trajectory metrics use the within-seed intersection of
episodes solved by all methods in each comparison. Cohort sizes are given in
the corresponding captions. This distinction prevents a lower-success method
from appearing efficient by solving only easier queries.

\subsection{Comparison With Baseline}
\label{sec:main-results}

\begin{figure*}[!t]
\centering
\includegraphics[width=0.99\textwidth]{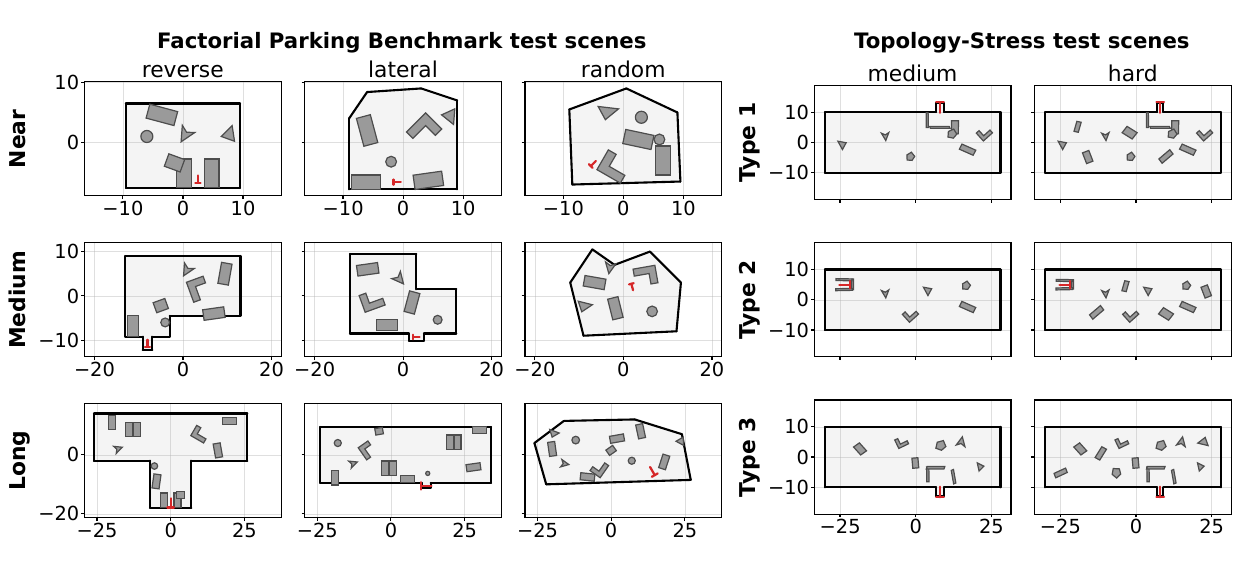}
\caption{Representative held-out scenes from the Factorial Parking Benchmark
(left) and the Topology-Stress Benchmark (right). The Factorial block arranges
start-separation regimes by row and target-pose types by column; the
Topology-Stress block arranges route-choice families by row and difficulty by
column. All panels preserve equal $x$--$y$ metric scaling. Red T markers denote
target poses and headings, and axes are in meters.}
\label{fig:sceneexamples}
\end{figure*}

We first illustrate the two evaluation settings with representative held-out
scenes from both benchmarks (Fig.~\ref{fig:sceneexamples}). Their construction
and splits are described in Section~\ref{sec:setup}, and the frozen evaluation
protocol is defined in Section~\ref{sec:method}.

\begin{figure}[!t]
\centering
\includegraphics[width=0.98\columnwidth]{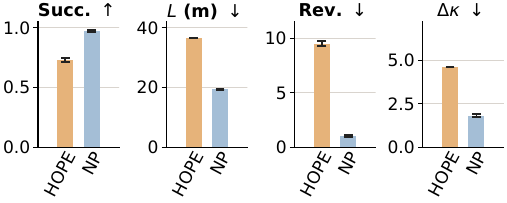}
\caption{NeuralParker improves success rate and trajectory quality over the
adapted HOPE planner on the Factorial Parking Benchmark. Success uses all 1080
episodes per seed; trajectory metrics use the pairwise common-success cohort
($n=773/748/790$). Bars and error bars show the three-seed mean $\pm$ sample
SD. NP denotes NeuralParker.}
\label{fig:main-comparison}
\vspace{-0.45\baselineskip}
\end{figure}

To assess complete-planner performance, we compare NeuralParker with the
adapted HOPE planner on the Factorial Parking Benchmark
(Fig.~\ref{fig:main-comparison}). NeuralParker achieves higher full-set success
and lower path length, reversal count, and cumulative curvature change on the
pairwise common-success cohort. The adapted HOPE baseline retains its native
action parameterization, learned-action curvature envelope, action mask, and RS
controller together with the attached fixed-Hermite terminal, so this
comparison is at the planner level.
Section~\ref{sec:hermite-results} separately isolates NeuralParker's terminal
connector on the same benchmark.

\begin{figure}[!t]
\centering
\includegraphics[width=0.98\columnwidth]{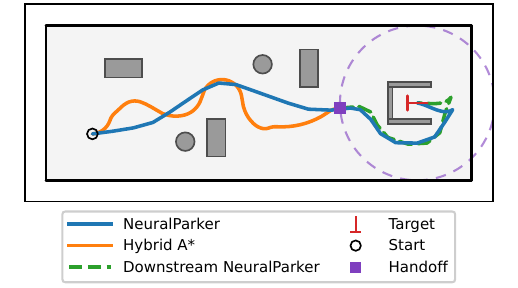}
\caption{Unified planning keeps one approach decision where a staged handoff
splits it. Each planner is drawn as one line, so the staged arm's arcs and
terminal connector share the dashed green line. One checkpoint drives both
learned arms on this pinned pocket case; the dashed circle is the
10 m handoff boundary.}
\label{fig:overview}
\end{figure}

We next test whether a classical long-range prefix can substitute for unified
planning by comparing NeuralParker with two staged decompositions
(Table~\ref{tab:twostage}). A region-goal Hybrid A* prefix drives the vehicle to
the first state within 10 m of the target and then hands control to a learned
planner. Both staged rows use the same prefix and continuous-curvature smoothing
when feasible, with a frozen raw-path fallback, and neither downstream policy
is retrained for the handoff. Unified planning gives the lowest path costs.
Handoff to the same NeuralParker policy retains a comparable success rate but
degrades all three trajectory metrics, whereas handoff to the adapted HOPE
planner also reduces success. Owning the whole approach lets the policy choose an approach
side and arrive in a state its terminal connector can finish cleanly, whereas a
geometrically reasonable prefix cut at a fixed radius can deliver the learned
stage to a pose that costs reversals to repair. Figure~\ref{fig:overview}
illustrates the two decompositions on one pinned case drawn from outside this
benchmark, because a three-sided pocket whose entrance faces away from the
arrival direction separates the two approach decisions more legibly than a
factorial scene does. Since that geometry belongs to the
route-choice families of Section~\ref{sec:representation-results}, both arms use
the target-relative vertices policy trained there, with one checkpoint driving
the unified arm and the staged arm's downstream stage.

\begin{table}[!t]
\centering
\caption{Unified Versus Staged Planning. Succ. uses all 1080 episodes
per seed; path metrics use the all-method common-success cohort
($n=700/719/726$). For staged rows, \dk sums the within-prefix and downstream
variations and excludes the single handoff curvature jump.}
\label{tab:twostage}
\footnotesize
\setlength{\tabcolsep}{0pt}
\begin{tabular*}{\columnwidth}{@{\extracolsep{\fill}}lcccc@{}}
\toprule
Method & Succ.~$\uparrow$ & $L$ (m)~$\downarrow$ &
Rev.~$\downarrow$ & \dk~$\downarrow$ \\
\midrule
\textbf{Unified NP} & \textbf{\msd{0.974}{0.007}} &
\textbf{\msd{19.77}{0.41}} & \textbf{\msd{1.02}{0.10}} &
\textbf{\msd{1.77}{0.10}} \\
Hybrid A* $\to$ NP & \msd{0.969}{0.002} & \msd{22.22}{0.50} &
\msd{1.42}{0.05} & \msd{3.05}{0.09} \\
Hybrid A* $\to$ HOPE & \msd{0.686}{0.014} & \msd{36.36}{0.38} &
\msd{6.97}{0.34} & \msd{5.16}{0.10} \\
\bottomrule
\end{tabular*}
\end{table}

\subsection{Global Geometry Ablation}
\label{sec:representation-results}

We next examine whether explicit global geometry helps when route selection
depends on structures outside the local LiDAR view. The Factorial
Parking Benchmark spans varied poses, distances, and clutter, and many of its
queries are locally resolvable: compact scenes expose most relevant geometry
to LiDAR, and many larger scenes permit a direct approach. We therefore conduct
this ablation on the separately designed Topology-Stress Benchmark, where the
target-side opening is initially outside the local view and the approach side
must be selected earlier.

To isolate the representation effect within each policy backbone, we compare
no added representation, BEV features, and target-relative vertices while
retaining each backbone's local state and 120-beam LiDAR input
(Table~\ref{tab:topology-stress}). All six configurations use the same geodesic
start curriculum and fixed-Hermite terminal connector. The three HOPE-style variants
additionally disable the 42-anchor action mask and the native Reeds--Shepp
takeover: the former samples an executed action from 42 anchors using the
policy density multiplied by LiDAR-derived safety weights, whereas the latter
overrides policy actions near the target. These rows therefore isolate scene
representation as controlled variants of the HOPE-style backbone, separately
from the complete adapted planner evaluated in
Section~\ref{sec:main-results}. Target-relative
vertices give the best aggregate result on every reported metric for both
backbones, with a more
pronounced advantage on the HOPE-style backbone. They also avoid the image
rendering and autoencoder pretraining required by the BEV input. Both backbones
retain their reported LiDAR preprocessing conventions, detailed in
Appendices~\ref{app:implementation} and~\ref{app:hope-adaptation}.

\begin{table}[!t]
\centering
\caption{Scene-Representation Ablation. Succ. uses all 720 episodes per seed;
path metrics use the six-method common-success cohort
($n=328/317/281$).}
\label{tab:topology-stress}
\footnotesize
\setlength{\tabcolsep}{2pt}
\begin{tabular*}{\columnwidth}{@{\extracolsep{\fill}}lcccc@{}}
\toprule
Added representation & Succ.~$\uparrow$ & $L$ (m)~$\downarrow$ &
Rev.~$\downarrow$ & \dk~$\downarrow$ \\
\midrule
\multicolumn{5}{@{}l}{\textit{NeuralParker backbone}} \\
none & \msd{0.897}{0.010} & \msd{37.84}{4.01} & \msd{2.96}{0.58} & \msd{5.74}{0.69} \\
BEV & \msd{0.848}{0.030} & \msd{38.72}{1.07} & \msd{4.41}{0.27} & \msd{6.25}{0.35} \\
vertices & \textbf{\msd{0.916}{0.023}} & \textbf{\msd{35.40}{0.93}} & \textbf{\msd{2.77}{0.10}} & \textbf{\msd{5.06}{0.55}} \\
\midrule
\multicolumn{5}{@{}l}{\textit{HOPE-style backbone}} \\
none & \msd{0.690}{0.018} & \msd{58.53}{1.54} & \msd{21.70}{1.03} & \msd{13.34}{0.63} \\
BEV & \msd{0.763}{0.100} & \msd{58.03}{2.61} & \msd{20.74}{4.21} & \msd{13.48}{1.54} \\
vertices & \textbf{\msd{0.808}{0.103}} & \textbf{\msd{49.79}{4.71}} & \textbf{\msd{16.95}{3.59}} & \textbf{\msd{10.85}{1.47}} \\
\bottomrule
\end{tabular*}
\end{table}

\begin{figure}[!t]
\centering
\includegraphics[width=0.98\columnwidth]{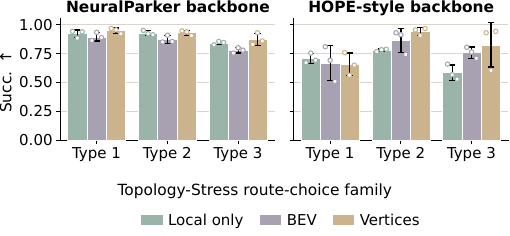}
\caption{Absolute family-level success rates show that representation effects
depend on route-choice family and policy backbone on the Topology-Stress
Benchmark. Each family contains 240 episodes per seed. Every group compares
local-only, BEV, and vertices; bars and error bars show the three-seed mean
$\pm$ sample SD, and open circles show individual seeds.}
\label{fig:global-family}
\end{figure}

We further disaggregate absolute success rates by route-choice family to test
whether the added representations help uniformly across scene geometries
(Fig.~\ref{fig:global-family}). With the NeuralParker backbone, vertices produce
small gains on Types 1 and 3 and little change on
Type 2, whereas BEV remains below the local-only configuration in all three
families. With the HOPE-style backbone, both added representations improve
Types 2 and 3 but not Type 1; several
added-representation variants also show appreciable between-seed variation. This
distinct response supports retaining Types 1
and 3 as separate families: the former constrains the
target-entry side, whereas the latter blocks an earlier outer approach and
exposes an offset passage. Each backbone keeps its own reported LiDAR
convention in this comparison: NeuralParker clips negative clearances to zero,
whereas the HOPE-style backbone retains them. Within-backbone contrasts
therefore carry the representation effect.

\subsection{Terminal Connector Ablation}
\label{sec:hermite-results}

To separate the immediate effect of enlarging the terminal candidate set from
the effect of training under the same rule, we compare three NeuralParker
connector configurations on the Factorial Parking Benchmark
(Table~\ref{tab:hermite}). The first configuration is the reference, using the
fixed nine-candidate Hermite connector. Substituting the
ensemble only at evaluation improves the success rate and all three path-quality
metrics without changing the policy. Training with the same ensemble retains
these gains and gives the highest success rate, the shortest paths, and the
lowest curvature variation, with a reversal count that matches the
evaluation-only replacement to within one seed standard deviation.

\begin{figure}[!t]
\centering
\includegraphics[width=0.98\columnwidth]{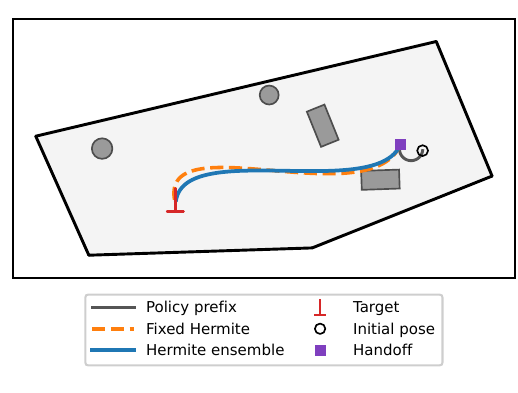}
\caption{Both terminal rules are feasible from the marked handoff pose, but the
ensemble selects a different terminal shape. Replaying only the connector from
a shared handoff isolates candidate selection from the policy prefix, which is
shown for context.}
\label{fig:hermite-case}
\end{figure}

To determine whether the ensemble improves terminal geometry even when the
fixed connector is already feasible, we replay both connectors from the same
policy handoff (Fig.~\ref{fig:hermite-case}). Both connectors are feasible from
that pose, so the difference comes from selecting a better terminal geometry.

\begin{table}[!t]
\centering
\caption{Terminal-Connector Ablation on the Factorial Parking Benchmark.
Succ. uses all 1080 episodes per seed; path metrics use the three-configuration
common-success cohort
($n=1003/1000/1015$).}
\label{tab:hermite}
\footnotesize
\setlength{\tabcolsep}{0pt}
\begin{tabular*}{\columnwidth}{@{\extracolsep{\fill}}lcccc@{}}
\toprule
Terminal rule & Succ.~$\uparrow$ & $L$ (m)~$\downarrow$ & Rev.~$\downarrow$ &
\dk~$\downarrow$ \\
\midrule
Fixed Hermite & \msd{0.965}{0.007} & \msd{20.13}{0.69} &
\msd{1.37}{0.09} & \msd{2.36}{0.15} \\
Ensemble, eval. only & \msd{0.972}{0.006} & \msd{18.76}{0.23} &
\textbf{\msd{1.11}{0.06}} & \msd{1.91}{0.07} \\
\textbf{Ensemble, train + eval.} & \textbf{\msd{0.974}{0.007}} &
\textbf{\msd{18.69}{0.11}} & \msd{1.15}{0.06} &
\textbf{\msd{1.86}{0.08}} \\
\bottomrule
\end{tabular*}
\end{table}

\subsection{Start-Curriculum Ablation}
\label{sec:curriculum-results}

We assess the role of start ordering by comparing geodesically ordered starts
with training that samples the full start range without a curriculum
(Table~\ref{tab:app-controls}). The two configurations share the
representation, reward, network, terminal connector, training budget, and
frozen full-range evaluation starts, so the comparison isolates the ordering.
Under the same fixed full-range evaluation, the two settings achieve
similar success, while geodesic ordering improves all three common-success
trajectory metrics. We therefore retain it as a training-recipe choice rather
than a separate method contribution.

\begin{table}[!t]
\centering
\caption{Start-Curriculum Ablation. Succ. uses all 1080 episodes per seed; path
metrics use the pairwise common-success cohort ($n=1043/1005/1028$). Values are
three-seed mean $\pm$ sample SD.}
\label{tab:app-controls}
\footnotesize
\setlength{\tabcolsep}{2pt}
\begin{tabular*}{\columnwidth}{@{\extracolsep{\fill}}lcccc@{}}
\toprule
Start curriculum & Succ.~$\uparrow$ & $L$ (m)~$\downarrow$ &
Rev.~$\downarrow$ & \dk~$\downarrow$ \\
\midrule
No curriculum & \msd{0.969}{0.015} & \msd{19.90}{0.75} &
\msd{1.43}{0.10} & \msd{2.23}{0.12} \\
Geodesic curriculum & \textbf{\msd{0.974}{0.007}} &
\textbf{\msd{18.76}{0.16}} & \textbf{\msd{1.17}{0.05}} &
\textbf{\msd{1.90}{0.06}} \\
\bottomrule
\end{tabular*}
\end{table}

\subsection{Real-Vehicle Experiments}
\label{sec:realworld}

\begin{figure*}[!t]
\centering
\subfloat[]{\includegraphics[width=0.245\textwidth,height=2.4cm]{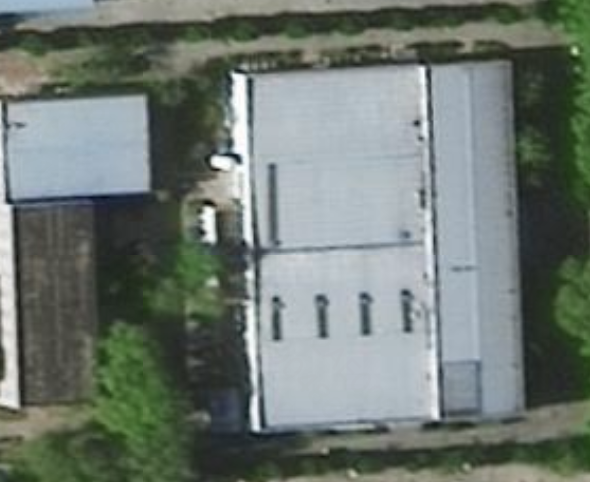}}\hfill
\subfloat[]{\includegraphics[width=0.245\textwidth,height=2.4cm]{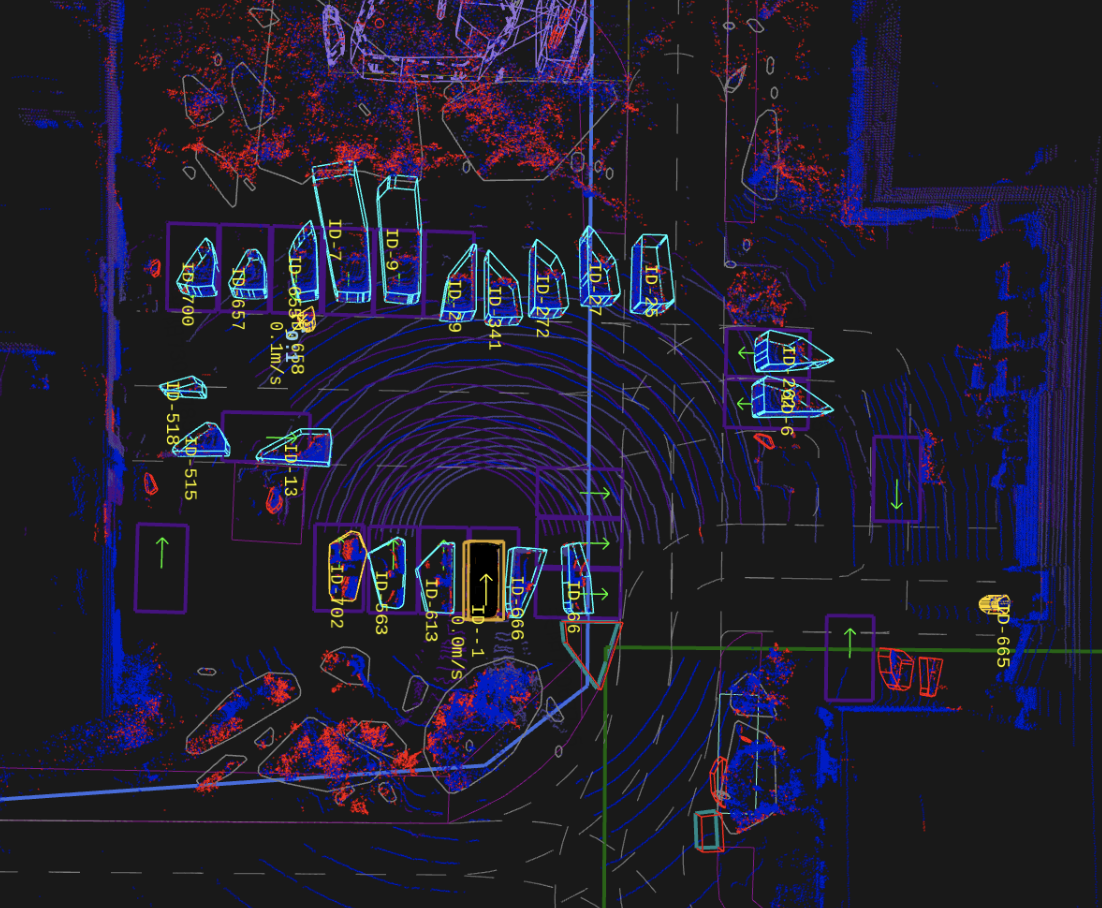}}\hfill
\subfloat[]{\includegraphics[width=0.245\textwidth,height=2.4cm]{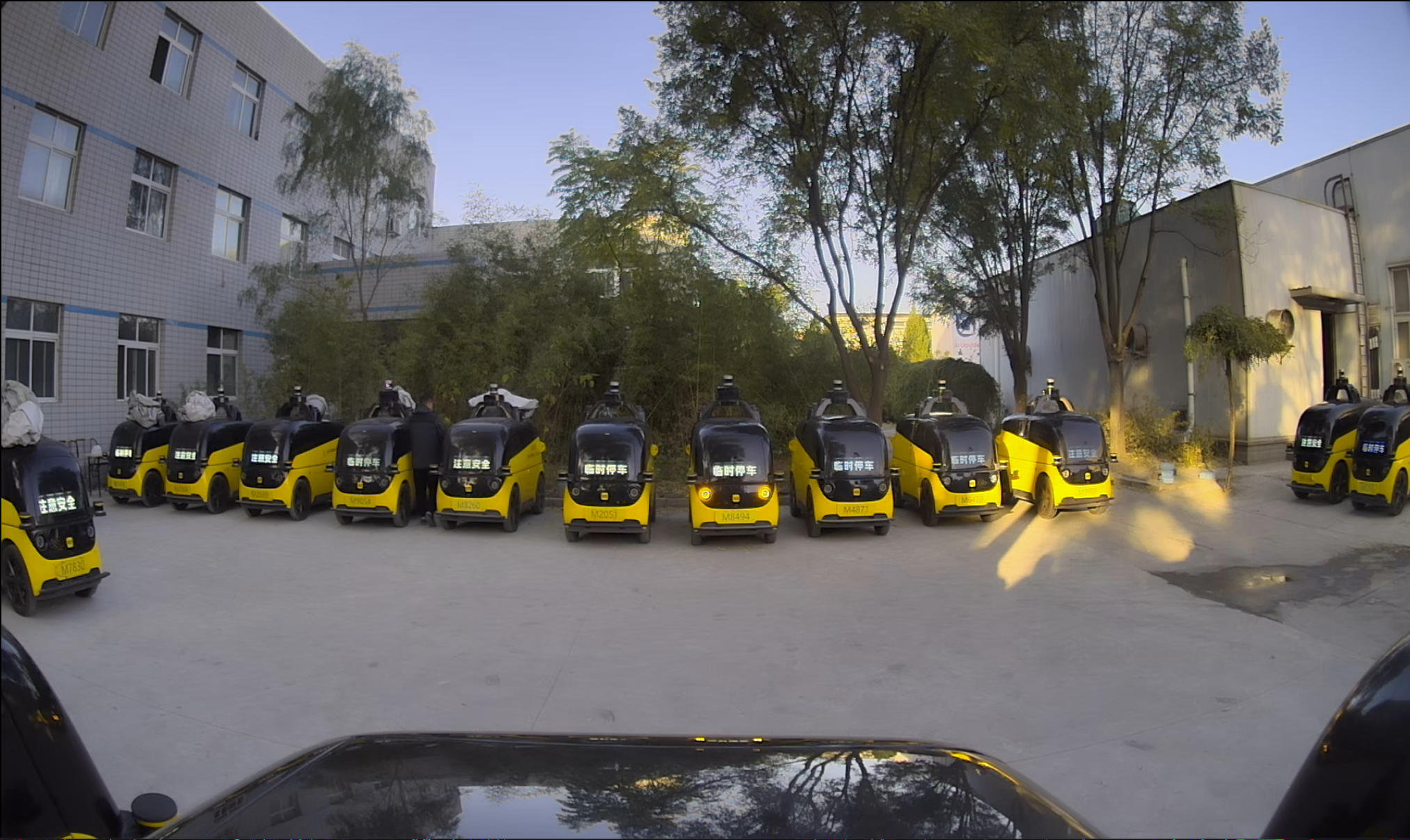}}\hfill
\subfloat[]{\includegraphics[width=0.245\textwidth,height=2.4cm]{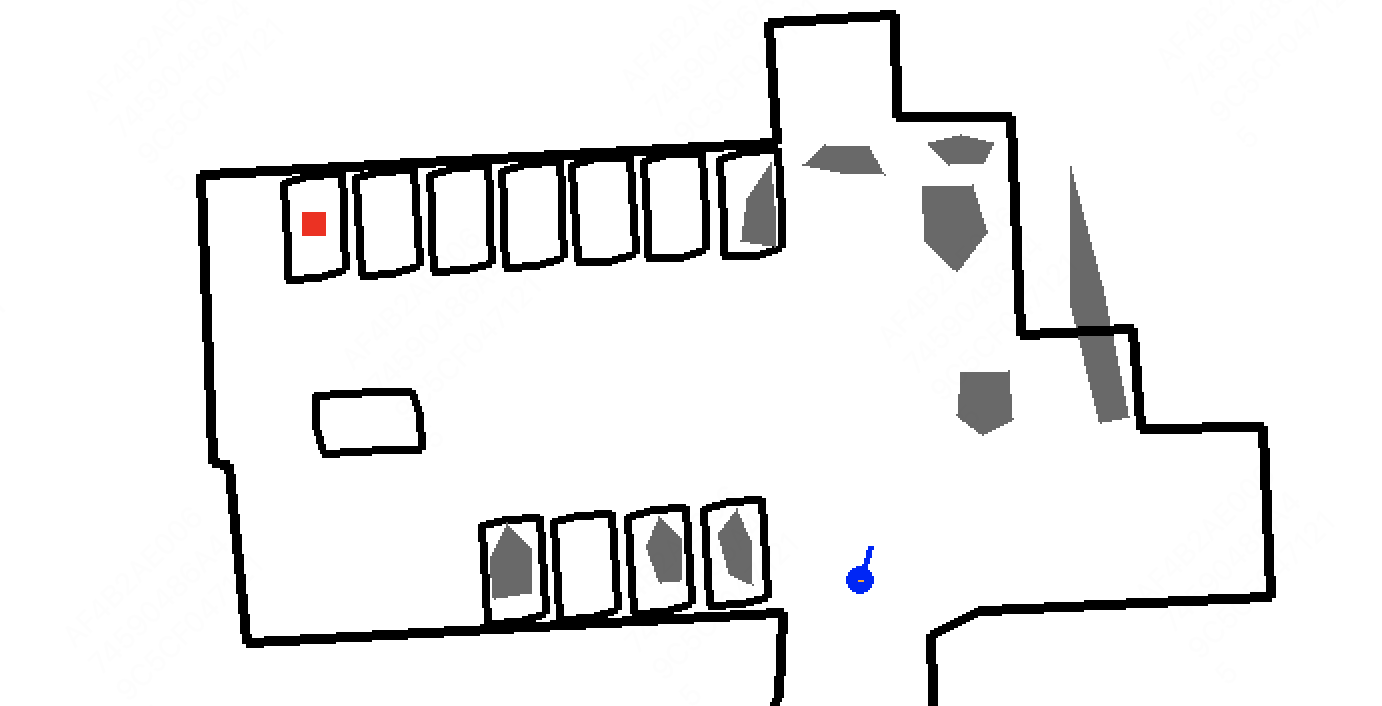}}\\[2pt]
\subfloat[]{\includegraphics[width=0.245\textwidth,height=2.4cm]{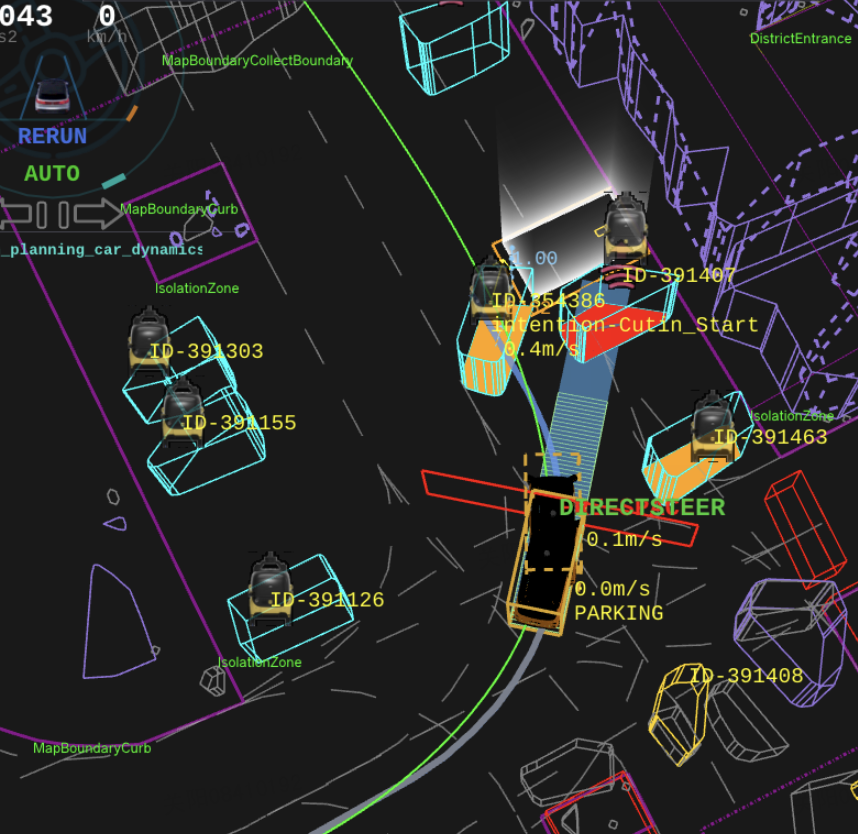}}\hfill
\subfloat[]{\includegraphics[width=0.245\textwidth,height=2.4cm]{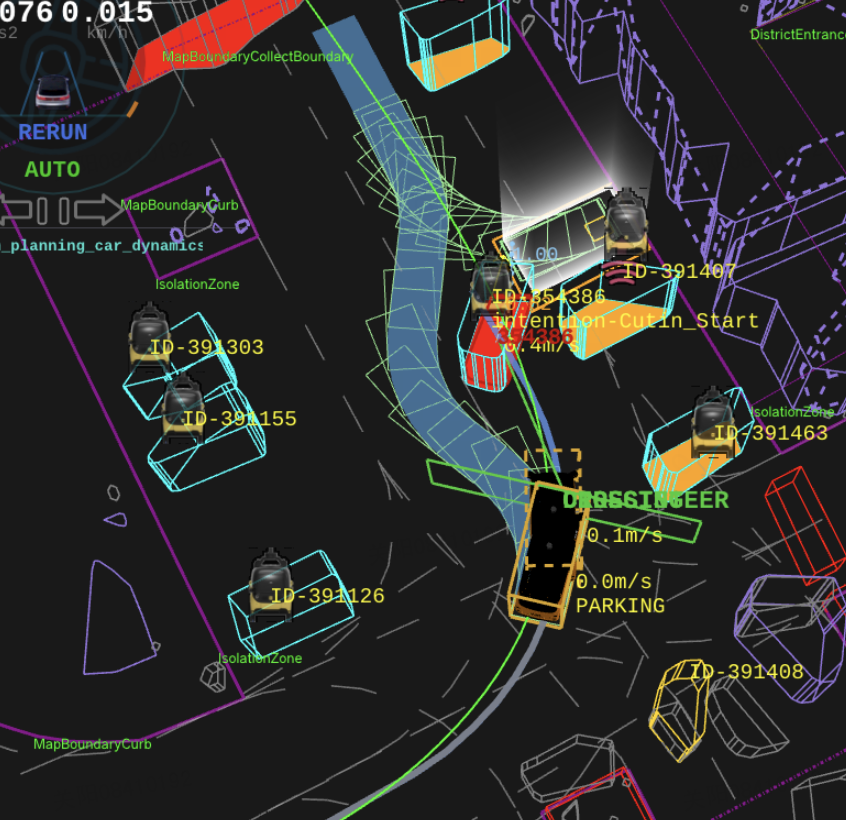}}\hfill
\subfloat[]{\includegraphics[width=0.245\textwidth,height=2.4cm]{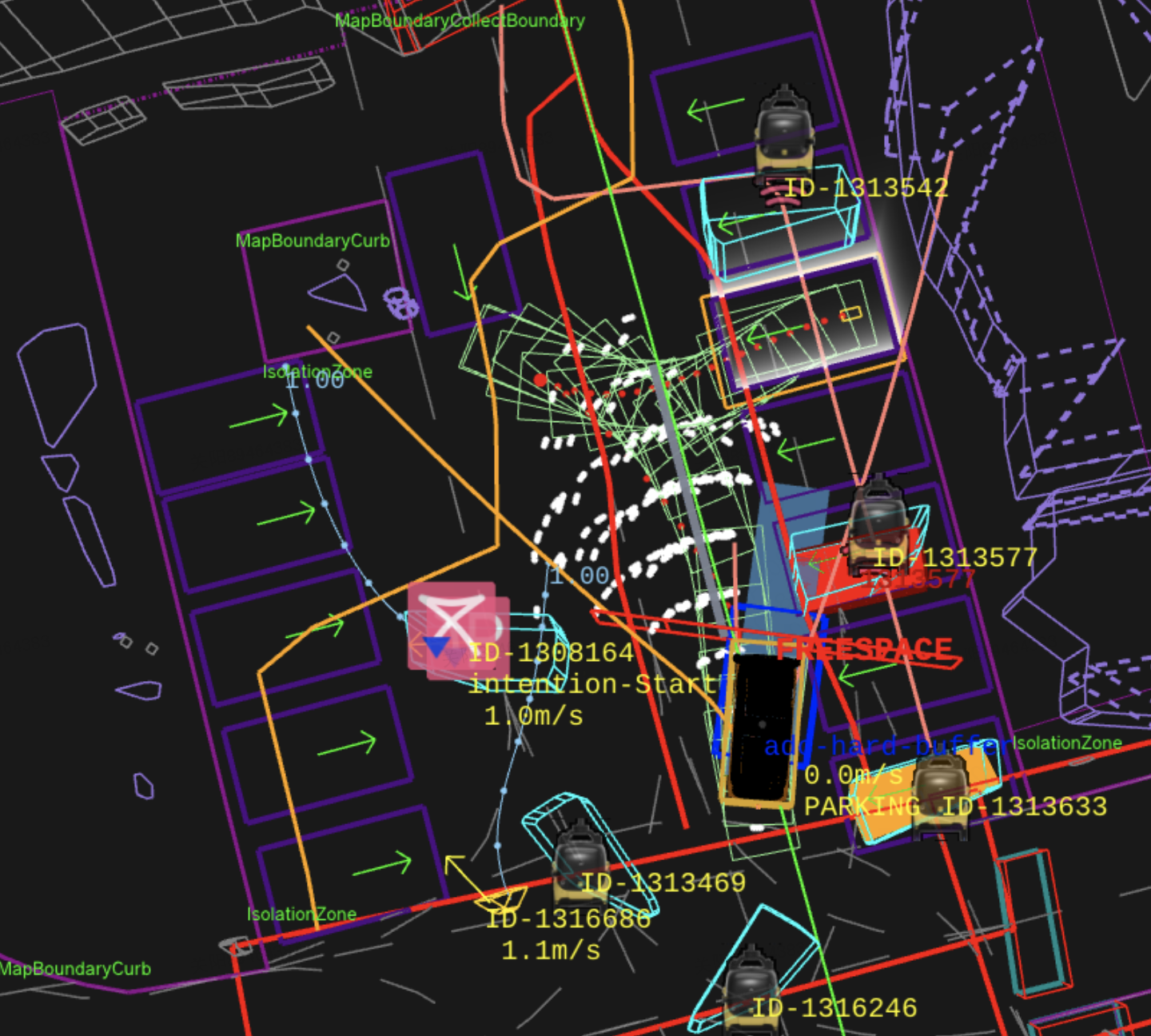}}\hfill
\subfloat[]{\includegraphics[width=0.245\textwidth,height=2.4cm]{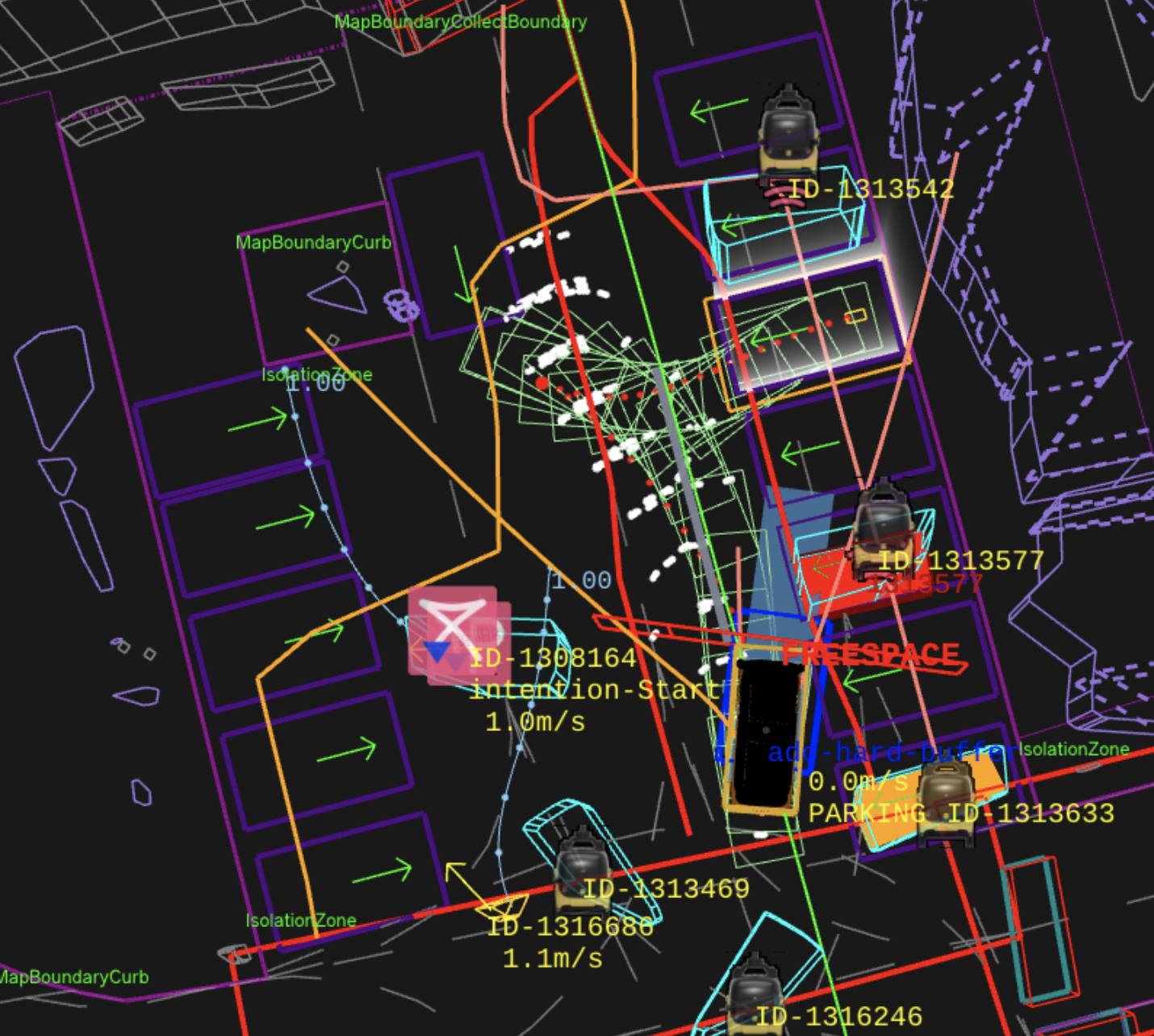}}
\caption{Real-vehicle experiments with a simpler NeuralParker configuration.
(a)--(d) operating site, perception visualization, camera view, and
site-specific simulation reconstruction; (e)--(h) two simulation-validated
cases comparing Hybrid A* with NeuralParker-Guided A*.}
\label{fig:realworld}
\end{figure*}

To assess planning under real-world perception, we evaluate a simpler
NeuralParker configuration on a real vehicle. It shares the target-relative
geometry representation and the arc policy, and uses the fixed nine-candidate
Hermite connector without the local LiDAR stream. The site is a
delivery-vehicle parking area in Hualikan, Beijing, shown in
Fig.~\ref{fig:realworld}. The planner consumes perception recorded by operating
delivery vehicles, the HD map, and an assigned parking pose, and replans from
these inputs at each frame. Its planned paths are then executed and validated
on the vehicle's simulation platform.

To compare the three planners under this protocol, we evaluate them on 47
difficult held-out clips using the validation rules summarized in
Table~\ref{tab:realworld}. Succ. is evaluated on all clips using
full-footprint validation, except for the parenthesized NeuralParker value
obtained with reference-point validation. The reported $L$ and Rev. values are
descriptive statistics computed on each method's own successful clips.
NeuralParker-Guided A* achieves the
highest full-footprint success, whereas direct NeuralParker has the lowest
planning time. Direct NeuralParker reaches 0.93 under the reference-point
criterion it was trained with, and the two criteria differ because training does
not model the full vehicle footprint.
Appendix~\ref{app:realworld-setup} reports how the learned guidance changes
Hybrid A* search time across these clips.

We further inspect two representative cases to clarify how learned guidance
changes the search behavior (Fig.~\ref{fig:realworld}(e)--(h)). Learned guidance
finds an indirect maneuver that first moves away from the target and reduces
the number of expanded nodes when both planners agree on the maneuver. The
site, operational data interface, and validation protocol are documented in
Appendix~\ref{app:realworld-setup}.

\begin{table}[!t]
\centering
\caption{Real-Vehicle Experiment on 47 Difficult Clips. Succ. uses all clips;
$L$ and Rev. are computed on each method's own successful clips.}
\label{tab:realworld}
\footnotesize
\setlength{\tabcolsep}{1.0pt}
\begin{tabular*}{\columnwidth}{@{\extracolsep{\fill}}lcccc@{}}
\toprule
Method & Succ.~$\uparrow$ & $L$ (m)~$\downarrow$ & Rev.~$\downarrow$ &
Plan time (ms)~$\downarrow$ \\
\midrule
NeuralParker & 0.60 (0.93) & 7.0 & 1.20 & \textbf{0.37} \\
NP-Guided A* & \textbf{0.72} & 7.7 & 1.17 & 4.24 \\
Hybrid A* & 0.67 & 7.2 & 1.13 & 5.26 \\
\bottomrule
\end{tabular*}
\end{table}


\section{Conclusion}
\label{sec:conclusion}

This work has presented NeuralParker, a reinforcement learning-based hybrid planner
for arbitrary-pose parking in irregular environments. It combines
target-relative full-environment vertices and local clearance rays with a
learned curvature--length arc policy and an in-loop, curvature-regularized
Hermite terminal ensemble. Factorial and topology-stress
benchmarks separate full-set success from trajectory quality on common-success
episodes.

Target-relative vertices provide the strongest aggregate success rate and
trajectory quality under both evaluated policy backbones, without the rendering
and pretraining required by the bird's-eye-view alternative. Replacing the
fixed Hermite connector with the ensemble improves the success rate and
trajectory quality without policy retraining, and training under the same terminal rule
retains these gains. At the planner level, NeuralParker achieves higher full-set
success and lower common-success trajectory costs than the adapted HOPE
planner, while unified planning gives better trajectory quality than the tested
staged decomposition. The real-vehicle evaluation takes the planner to an
operating delivery site, where it plans from perception recorded by real
delivery vehicles and the HD map, and its paths are validated in simulation.

The reported results use collision checks on the vehicle reference point,
static procedurally constructed 2-D benchmarks with fixed object budgets, and
three policy-training seeds. Dynamic actors, perception uncertainty, broader
object counts and topologies, and closed-loop validation of the
vertices-plus-LiDAR planner remain for future work.
Overall, the results show that retaining target-side geometry and
coupling policy learning to a diverse terminal connector extends hybrid RL
parking from local slot maneuvers to longer-range arbitrary-pose planning.


\appendices

\section{NeuralParker Implementation Details}
\label{app:implementation}

To support reproducibility of the reported NeuralParker results, we document
the observation, network, planner, and training settings used in the main
simulation experiments. These parameters describe the vertices-plus-LiDAR
policy with the 81-candidate Hermite terminal ensemble.

\subsection{Policy and Planner Configuration}

To specify the evaluated policy and planner, we summarize the observation,
network, learned-action, and terminal-connector settings in
Table~\ref{tab:app-implementation}.

\begin{table*}[!t]
\centering
\caption{NeuralParker Configuration}
\label{tab:app-implementation}
\footnotesize
\begin{minipage}[t]{0.485\textwidth}
\centering
\textit{Panel A: Observation and network}\par\vspace{2pt}
\setlength{\tabcolsep}{3pt}
\begin{tabularx}{\linewidth}{@{}p{2.35cm}X@{}}
\toprule
Parameter & Setting \\
\midrule
Coordinate frame & Target at $(0,0)$ with heading $+y$ \\
Obstacle capacity & $20$ slots $\times$ 16 values \\
Boundary capacity & $10$ slots $\times$ 4 values \\
Local LiDAR & 120 rectified footprint-clearance beams, $360^\circ$,
10 m ray cap \\
Geometry tokens & 16-D, one attention head \\
Global--local fusion & 128-D, depth 1, four 32-D heads \\
Actor / critic MLP & Separate $[64,128,256,64]$ networks \\
\bottomrule
\end{tabularx}
\end{minipage}\hfill
\begin{minipage}[t]{0.485\textwidth}
\centering
\textit{Panel B: Learned action and terminal connector}\par\vspace{2pt}
\setlength{\tabcolsep}{3pt}
\begin{tabularx}{\linewidth}{@{}p{2.35cm}X@{}}
\toprule
Parameter & Setting \\
\midrule
Learned action & $\kappa\in[-0.68,0.68]$,
$l\in[-10,10]$ m \\
Episode / tolerance & 200 arcs; 0.1 m and $10^\circ$ \\
Endpoint grid & offsets $\{-0.1,0,0.1\}$ m; headings
$\{80^\circ,90^\circ,100^\circ\}$ \\
Tangent scales & independent $\{0.8,1.0,1.2\}^2$ \\
Candidate count & $9\times3\times3=81$ \\
Heading gate & $\sin\tilde\theta>0$ at the reached state \\
Feasibility samples & 100 points; $\bar\kappa\leq0.68$ \\
Selection weight & $c_{\mathrm{sel}}=10$ in~\eqref{eq:mincost} \\
Terminal-score floor & $p_H\geq-500$ for the ensemble \\
\bottomrule
\end{tabularx}
\end{minipage}
\end{table*}

To fully specify the observation and collision-checking pipeline, we record
how unused slots, LiDAR clearances, and path samples are handled. Unused
geometry slots contain the $-10^4$ sentinel and are not explicitly
attention-masked. NeuralParker uses the rectified clearance
$s_i^\ell=\max(d_i-b_i,0)$ without observation normalization; empty beams
therefore retain the direction-dependent value $10-b_i$, and circular obstacles
use 16-segment rings for ray casting. For the reported simulation runs, $d_i$
is produced by the benchmark ray caster using strict, zero-tolerance
segment-bound tests; training and evaluation use the same caster. Learned arcs
use 30 sampled reference points for the hard collision and boundary checks;
Hermite candidates use the 100 samples listed in
Table~\ref{tab:app-implementation}.

\subsection{Training Configuration}

To make NeuralParker training and checkpoint selection reproducible, we report
the complete training recipe in Table~\ref{tab:app-ppo}, where GAE denotes
generalized advantage estimation.
Snapshots are saved every 20 epochs, and the reported checkpoints are selected
by per-epoch mean reward over 32 sampled full-range episodes from the
training-scene distribution. The selected epochs are 227, 182, and 228 for
curriculum-enabled seeds 0--2 and 234, 148, and 234 for their matched
no-curriculum controls. No frozen evaluation episode is used for checkpoint
selection.

\begin{table}[!t]
\centering
\caption{NeuralParker Training Configuration}
\label{tab:app-ppo}
\footnotesize
\setlength{\tabcolsep}{4pt}
\begin{tabular*}{\columnwidth}{@{\extracolsep{\fill}}lr@{}}
\toprule
Parameter & Setting \\
\midrule
Learning rate / schedule & $3\times10^{-4}$ / linear decay \\
Discount $\gamma$ / GAE $\lambda$ & 0.99 / 0.95 \\
PPO clip / value coefficient & 0.15 / 0.25 \\
Entropy coefficient / gradient norm & 0.005 / 0.5 \\
Advantage / reward normalization & yes / yes \\
Epochs / steps per epoch & 240 / 60,000 \\
Steps per collect / repeats & 16,384 / 10 \\
Batch / replay-buffer size & 8,192 / 131,072 \\
Training / test environments & 128 / 32 \\
Curriculum start scale / distance & 0.05 / geodesic \\
\bottomrule
\end{tabular*}
\end{table}

\section{Adaptation of HOPE to the Evaluation Benchmarks}
\label{app:hope-adaptation}

We adapt HOPE for two purposes: to obtain a complete planner baseline under the
common task definition and to construct a HOPE-style backbone for the
scene-representation ablation. We also reuse the complete baseline in the
staged diagnostic. The following subsections separate the learning contract,
benchmark interface, and controlled representation inputs.

\subsection{Policy and Training Contract}

We preserve HOPE's published actor--critic architecture, PPO implementation,
reward, and action parameterization in the complete adapted baseline. Each seed
is trained for 10,000 episodes from full-range starts without a start
curriculum, and the checkpoint saved after 10,000 episodes is evaluated. The baseline retains the
native 42-anchor LiDAR action mask and Reeds--Shepp (RS) controller and uses the
attached fixed nine-candidate Hermite terminator. The staged diagnostic
performs no retraining; it reuses the corresponding complete-baseline
checkpoint and downstream configuration.

We retain HOPE's attention-fusion architecture and optimization settings while
changing only the active input-modality branches required by the controlled
representations.
Each active modality is embedded in 128-D, and one attention block fuses the
embeddings using eight 32-D heads and a 128-unit feed-forward sublayer; a
128-unit hidden layer follows the flattened output. The actor and critic
learning rates are $5\times10^{-6}$ and $2.5\times10^{-5}$,
respectively, with $\gamma=0.98$, GAE $\lambda=0.95$, clip ratio 0.2, batch
size 8192, mini-batch size 32, and ten update epochs. Advantages are normalized
and rewards are not. Running mean-and-standard-deviation normalization is
applied only to the LiDAR, target, and optional vertex branches, not to the
image or action mask.

\subsection{Benchmark Interface and Analytic Control}

We align HOPE with the common task definition by replacing its
collision-and-retreat behavior with hard reference-point collision and boundary termination.
All adapted-HOPE policies use the benchmark ray distances $d_i$, but retain
signed footprint clearance $d_i-b_i$ followed by per-beam running
mean-and-standard-deviation normalization. The
training statistics are stored with each checkpoint and restored for
evaluation. Thus, negative clearances are retained, whereas NeuralParker
rectifies them as described in Appendix~\ref{app:implementation}. The complete
baseline still uses rectangular-footprint geometry in its action mask, shaping
reward, and BEV rendering.

We preserve HOPE's learned-action curvature bound of
$\tan(0.75)/2.8\approx0.333~\mathrm{m}^{-1}$ and apply the common
$0.68~\mathrm{m}^{-1}$ feasibility cap only to the attached fixed-Hermite
terminator. RS candidates are generated and checked at 0.1 m spacing; learned
arcs and accepted RS chunks use 30 sampled reference points for the hard
checks, and Hermite candidates use 100. Accepted RS chunks override the policy
output but still pass through the shared environment, so hard termination and
the fixed-Hermite test remain active during RS takeover.

\subsection{Controlled Representation Inputs}

We construct the HOPE-style representation configurations with a shared
target-and-signed-LiDAR base and three choices for the added scene input: none, a BEV raster, and
target-relative vertices. Each configuration retains HOPE's PPO, reward, and action
parameterization, is trained for 10,000 episodes per seed with the geodesic
start curriculum at scale 0.05, and uses the checkpoint saved after 10,000
episodes. All three configurations
disable the action mask and RS controller but retain the fixed nine-candidate
Hermite terminator, which keeps the added scene input as the only difference
between them. The vertex configuration uses the attention-based vertex input
branch and no image encoder.

We implement both policy backbones' BEV controls with the published HOPE image-processing
path and a frozen convolutional autoencoder. The HOPE-style control retains a
trajectory tail of up to 20 poses, whereas the NeuralParker-backbone control
disables it. The renderer produces an ego-aligned $64\times64$ RGB raster from
target-frame geometry, the destination, and the ego footprint. Each autoencoder
is pretrained on 50,000 tail-enabled images from random-policy rollouts, using
4- and 8-channel convolutional stages, a 256-D fully connected layer, a 128-D
latent, Adam at $10^{-3}$, mean-squared-error loss, batch size 256, and 50
epochs. Separate
autoencoders are trained for the Factorial Parking and Topology-Stress training
distributions, and their encoders remain frozen during PPO.

\FloatBarrier

\section{Real-Vehicle Experiments}
\label{app:realworld-setup}

We document the operational inputs, simulation-based validation, and planner
configuration of the real-vehicle evaluation reported in
Section~\ref{sec:realworld}. The evaluation uses perception recorded during
delivery-vehicle operation together with the HD map and an assigned parking
pose, and planned paths are executed and validated in simulation. The simpler
planner configuration uses processed polygon and map geometry without
policy-level LiDAR and a fixed-tangent, nine-pose Hermite connector.

To reconstruct the operational site for training and held-out evaluation, we
use records collected in a delivery-vehicle parking area in Hualikan, Beijing.
Automated delivery vehicles return to a parking spot there after completing a
delivery. Static boundaries were extracted from the HD map as 50 polylines,
and a site-specific training simulator was reconstructed from 1,000
operational clips containing 95,000 obstacle-detection frames. At each
training reset, one recorded obstacle state was sampled, the nearest ten
detections were represented as five-vertex polygons, and valid ego and target
poses were generated. The 47 difficult evaluation clips were excluded from
this training set; dedicated validators then assessed task completion and
collision with obstacles or boundaries.

The three planners run as compiled C++ binaries in the vehicle software system.
Fig.~\ref{fig:realworld} visualizes LiDAR detections, whereas the evaluated
policy receives target-relative ego, obstacle-polygon, and map-boundary
geometry. During evaluation, it replans once per recorded frame from this
geometry, the HD map, and the assigned target. Each path is then executed in
simulation and checked under both the reference-point convention inherited from
training and a full rectangular vehicle-footprint validator. Reversals follow
\eqref{eq:metrics}.

\begin{figure}[!t]
\centering
\includegraphics[width=0.98\columnwidth]{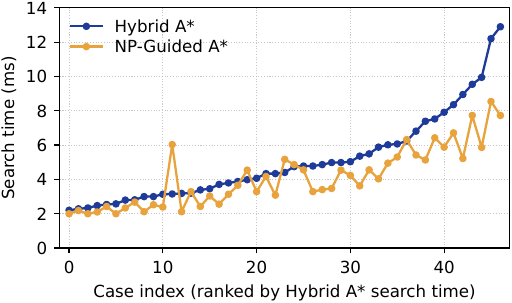}
\caption{Learned guidance reduces Hybrid A* search time mostly on the harder
real-vehicle clips. The 47 held-out clips are ordered by Hybrid A* search
time, which serves as a difficulty proxy, and both planners' times are shown in
that order.}
\label{fig:searchtime}
\end{figure}

To characterize how the learned guidance affects search effort, we rank the 47
clips by Hybrid A* search time and plot both planners' per-clip times in that
order (Fig.~\ref{fig:searchtime}). The ranking serves as a proxy for clip
difficulty. The two planners are comparable on the easier clips, whereas
NeuralParker-Guided A* increasingly undercuts Hybrid A* as the ranked search
time grows, so the saving concentrates on the harder queries. Individual clips
exceed the Hybrid A* time when the learned reference path disagrees with the
search's preferred maneuver.

\begin{thebibliography}{99}

\bibitem{paden2016}
B. Paden, M. \v{C}\'ap, S. Z. Yong, D. Yershov, and E. Frazzoli, ``A survey of
motion planning and control techniques for self-driving urban vehicles,''
\emph{IEEE Trans. Intell. Veh.}, vol. 1, no. 1, pp. 33--55, 2016.

\bibitem{banzhaf2017}
H. Banzhaf, D. Nienh\"user, S. Knoop, and J. M. Z\"ollner, ``The future of
parking: A survey on automated valet parking with an outlook on high density
parking,'' in \emph{Proc. IEEE Intell. Veh. Symp. (IV)}, 2017, pp. 1827--1834.

\bibitem{freightpose2021}
K. Masood, D. P. Morales, V. Fremont, M. Zoppi, and R. Molfino, ``Parking pose
generation for autonomous freight collection by pallet handling car-like
robot,'' \emph{Energies}, vol. 14, no. 15, art. no. 4677, 2021.

\bibitem{dubins1957}
L. E. Dubins, ``On curves of minimal length with a constraint on average
curvature, and with prescribed initial and terminal positions and tangents,''
\emph{Amer. J. Math.}, vol. 79, no. 3, pp. 497--516, 1957.

\bibitem{reedsshepp1990}
J. A. Reeds and L. A. Shepp, ``Optimal paths for a car that goes both forwards
and backwards,'' \emph{Pacific J. Math.}, vol. 145, no. 2, pp. 367--393, 1990.

\bibitem{pivtoraiko2009}
M. Pivtoraiko, R. A. Knepper, and A. Kelly, ``Differentially constrained mobile
robot motion planning in state lattices,'' \emph{J. Field Robot.}, vol. 26,
no. 3, pp. 308--333, 2009.

\bibitem{dolgov2010}
D. Dolgov, S. Thrun, M. Montemerlo, and J. Diebel, ``Path planning for
autonomous vehicles in unknown semi-structured environments,'' \emph{Int. J.
Robot. Res.}, vol. 29, no. 5, pp. 485--501, 2010.

\bibitem{optca2018}
X. Zhang, A. Liniger, A. Sakai, and F. Borrelli, ``Autonomous parking using
optimization-based collision avoidance,'' in \emph{Proc. IEEE Conf. Decis.
Control (CDC)}, 2018, pp. 4327--4332.

\bibitem{optiter2021}
B. Li \emph{et al.}, ``Optimization-based trajectory planning for autonomous
parking with irregularly placed obstacles: A lightweight iterative framework,''
\emph{IEEE Trans. Intell. Transp. Syst.}, vol. 23, no. 8, pp. 11970--11981,
2022.

\bibitem{dai2021}
S. Dai and Y. Wang, ``Long-horizon motion planning for autonomous vehicle
parking incorporating incomplete map information,'' in \emph{Proc. IEEE Int.
Conf. Robot. Autom. (ICRA)}, 2021, pp. 8135--8142.

\bibitem{rle2e2019}
P. Zhang \emph{et al.}, ``Reinforcement learning-based end-to-end parking for
automatic parking system,'' \emph{Sensors}, vol. 19, no. 18, art. no. 3996,
2019.

\bibitem{rlmp2020}
J. Zhang, H. Chen, S. Song, and F. Hu, ``Reinforcement learning-based motion
planning for automatic parking system,'' \emph{IEEE Access}, vol. 8, pp.
154485--154501, 2020.

\bibitem{parkbench2026}
F. Tao \emph{et al.}, ``Adapting reinforcement learning for path planning in
constrained parking scenarios,'' arXiv:2601.22545, 2026.

\bibitem{parkinge2e2024}
C. Li, Z. Ji, Z. Chen, T. Qin, and M. Yang, ``ParkingE2E: Camera-based
end-to-end parking network, from images to planning,'' in \emph{Proc. IEEE/RSJ
Int. Conf. Intell. Robots Syst. (IROS)}, 2024, pp. 13206--13212.

\bibitem{parkformer2025}
J. Fu, B. Tian, H. Chen, S. Meng, and T. Yao, ``ParkFormer: A transformer-based
parking policy with goal embedding and pedestrian-aware control,'' in
\emph{Proc. IEEE Int. Intell. Transp. Syst. Conf. (ITSC)}, 2025,
pp. 4317--4323.

\bibitem{multipark2025}
H. Zheng \emph{et al.}, ``MultiPark: Multimodal parking transformer with
next-segment prediction,'' arXiv:2508.11537, 2025.

\bibitem{hope2024}
M. Jiang, Y. Li, S. Zhang, S. Chen, C. Wang, and M. Yang, ``HOPE: A
reinforcement learning-based hybrid policy path planner for diverse parking
scenarios,'' \emph{IEEE Trans. Intell. Transp. Syst.}, vol. 26, no. 5, pp.
6130--6141, 2025.

\bibitem{rlogm2025}
Z. Wang, Z. Chen, M. Jiang, T. Qin, and M. Yang, ``RL-OGM-Parking: LiDAR
OGM-based hybrid reinforcement learning planner for autonomous parking,'' in
\emph{Proc. IEEE Int. Conf. Robot. Autom. (ICRA)}, 2025, pp. 8420--8426.

\bibitem{n3p2026}
Y. Xue \emph{et al.}, ``N3P: Accelerated automated parking via a
learning-based naturalistic three-stage scheme,'' arXiv:2605.22722, 2026.

\bibitem{drip2025}
M. Jiang, Y. Li, J. Zhang, S. Zhang, and M. Yang, ``A diffusion-refined planner
with reinforcement learning priors for confined-space parking,''
arXiv:2510.14000, 2025.

\bibitem{vorobieva2015}
H. Vorobieva, S. Glaser, N. Minoiu-Enache, and S. Mammar, ``Automatic parallel
parking in tiny spots: Path planning and control,'' \emph{IEEE Trans. Intell.
Transp. Syst.}, vol. 16, no. 1, pp. 396--410, 2015.

\bibitem{upadhyay2018}
S. Upadhyay and A. Ratnoo, ``A point-to-ray framework for generating smooth
parallel parking maneuvers,'' \emph{IEEE Robot. Autom. Lett.}, vol. 3, no. 2,
pp. 1268--1275, 2018.

\bibitem{tazaki2017}
Y. Tazaki, H. Okuda, and T. Suzuki, ``Parking trajectory planning using
multiresolution state roadmaps,'' \emph{IEEE Trans. Intell. Veh.}, vol. 2,
no. 4, pp. 298--307, 2017.

\bibitem{guidedha2019}
S. Sedighi, D.-V. Nguyen, and K.-D. Kuhnert, ``Guided hybrid A-star path
planning algorithm for valet parking applications,'' in \emph{Proc. 5th Int.
Conf. Control, Autom. Robot. (ICCAR)}, 2019, pp. 570--575.

\bibitem{fastastar2021}
J. He and H. Li, ``Fast A* anchor point based path planning for narrow space
parking,'' in \emph{Proc. IEEE Int. Intell. Transp. Syst. Conf. (ITSC)}, 2021,
pp. 1604--1609.

\bibitem{rrtpark2018}
K. Zheng and S. Liu, ``RRT based path planning for autonomous parking of
vehicle,'' in \emph{Proc. IEEE 7th Data Driven Control Learn. Syst. Conf.
(DDCLS)}, 2018, pp. 627--632.

\bibitem{rrtavp2021}
S. Solmaz, R. Muminovic, A. Civgin, and G. Stettinger, ``Development, analysis,
and real-life benchmarking of RRT-based path planning algorithms for automated
valet parking,'' in \emph{Proc. IEEE Int. Intell. Transp. Syst. Conf. (ITSC)},
2021, pp. 621--628.

\bibitem{li2015unified}
B. Li and Z. Shao, ``A unified motion planning method for parking an autonomous
vehicle in the presence of irregularly placed obstacles,''
\emph{Knowl.-Based Syst.}, vol. 86, pp. 11--20, 2015.

\bibitem{reachable2023}
I. H. Oh, J. W. Seo, J. S. Kim, and C. C. Chung, ``Reachable set-based path
planning for automated vertical parking system,'' in \emph{Proc. IEEE 26th
Int. Conf. Intell. Transp. Syst. (ITSC)}, 2023, pp. 1194--1200.

\bibitem{guan2023idc}
Y. Guan \emph{et al.}, ``Integrated decision and control: Toward interpretable
and computationally efficient driving intelligence,'' \emph{IEEE Trans.
Cybern.}, vol. 53, no. 2, pp. 859--873, 2023.

\bibitem{guan2026enhancedidc}
Y. Guan \emph{et al.}, ``Enhanced integrated decision and control for
high-level automated vehicles and its experiment verification,'' \emph{IEEE
Trans. Autom. Sci. Eng.}, vol. 23, pp. 11578--11595, 2026.

\bibitem{drltraj2020}
Z. Du, Q. Miao, and C. Zong, ``Trajectory planning for automated parking systems
using deep reinforcement learning,'' \emph{Int. J. Automot. Technol.}, vol. 21,
no. 4, pp. 881--887, 2020.

\bibitem{revpark2019}
E. Bejar and A. Moran, ``Reverse parking a car-like mobile robot with deep
reinforcement learning and preview control,'' in \emph{Proc. IEEE 9th Annu.
Comput. Commun. Workshop Conf. (CCWC)}, 2019, pp. 377--383.

\bibitem{fedpark2023}
Z. Yuan, Z. Wang, X. Li, L. Li, and L. Zhang, ``Hierarchical trajectory planning
for narrow-space automated parking with deep reinforcement learning: A
federated learning scheme,'' \emph{Sensors}, vol. 23, no. 8, art. no. 4087,
2023.

\bibitem{rlexpert2023}
Y. Wu, L. Wang, X. Lu, Y. Wu, and H. Zhang, ``Reinforcement learning-based
autonomous parking with expert demonstrations,'' in \emph{Proc. 7th CAA Int.
Conf. Veh. Control Intell. (CVCI)}, 2023, pp. 1--6.

\bibitem{humanrev2024}
Z. Qiu, S. Chen, J. Shi, F. Wang, and N. Zheng, ``Human-like reverse parking
using deep reinforcement learning with attention mechanism,'' in \emph{Proc.
IEEE Intell. Veh. Symp. (IV)}, 2024, pp. 2553--2560.

\bibitem{transparking2025}
H. Du and C.-M. Chew, ``TransParking: A dual-decoder transformer framework with
soft localization for end-to-end automatic parking,'' arXiv:2503.06071, 2025.

\bibitem{vectornet2020}
J. Gao \emph{et al.}, ``VectorNet: Encoding HD maps and agent dynamics from
vectorized representation,'' in \emph{Proc. IEEE/CVF Conf. Comput. Vis.
Pattern Recognit. (CVPR)}, 2020, pp. 11522--11530.

\bibitem{vad2023}
B. Jiang \emph{et al.}, ``VAD: Vectorized scene representation for efficient
autonomous driving,'' in \emph{Proc. IEEE/CVF Int. Conf. Comput. Vis. (ICCV)},
2023, pp. 8340--8350.

\bibitem{guan2021directindirect}
Y. Guan \emph{et al.}, ``Direct and indirect reinforcement learning,''
\emph{Int. J. Intell. Syst.}, vol. 36, no. 8, pp. 4439--4467, 2021.

\bibitem{ppo2017}
J. Schulman, F. Wolski, P. Dhariwal, A. Radford, and O. Klimov, ``Proximal
policy optimization algorithms,'' arXiv:1707.06347, 2017.

\bibitem{vaswani2017}
A. Vaswani \emph{et al.}, ``Attention is all you need,'' in \emph{Proc. Adv.
Neural Inf. Process. Syst. (NeurIPS)}, vol. 30, 2017.

\bibitem{florensa2017}
C. Florensa, D. Held, M. Wulfmeier, M. Zhang, and P. Abbeel, ``Reverse
curriculum generation for reinforcement learning,'' in \emph{Proc. Conf. Robot
Learn. (CoRL)}, vol. 78, 2017, pp. 482--495.

\bibitem{barc2019}
B. Ivanovic, J. Harrison, A. Sharma, M. Chen, and M. Pavone, ``BaRC: Backward
reachability curriculum for robotic reinforcement learning,'' in \emph{Proc.
IEEE Int. Conf. Robot. Autom. (ICRA)}, 2019, pp. 15--21.

\end{thebibliography}
\end{document}